\documentclass[
dvipsnames, table,   %
format=sigconf,
anonymous=false,     %
review=false,        %
authorversion=false, %
]{acmart}

\usepackage{siunitx}

\usepackage{tikz}
\usetikzlibrary{arrows, decorations.markings}

\usepackage{xspace}

\usepackage{algcompatible, algorithm}
\usepackage[noend]{algpseudocode}

\newcommand{\best}{{\cellcolor[gray]{0.75}}}
\newcommand{\statsimilar}{{\cellcolor[gray]{0.9}}}

\newcommand\mX{\mathcal{X}}

\newcommand\Real{\mathbb{R}}
\newcommand\Normal{\mathcal{N}}

\newcommand{\bx}{\mathbf{x}}
\newcommand{\bI}{\mathbf{I}}
\newcommand{\xnext}{\bx'}
\newcommand{\bkappa}{\boldsymbol{\kappa}}
\newcommand{\btheta}{\boldsymbol{\theta}}
\newcommand{\fstar}{f^\star}
\newcommand{\fmin}{f_{\min}}

\newcommand{\given}{\,|\,}

\DeclareMathOperator*{\argmax}{\arg\!\max}
\DeclareMathOperator*{\argmin}{\arg\!\min}

\newcommand{\mGP}{\ensuremath{\mathcal{GP}}\xspace}

\newcommand{\Data}{\mathcal{D}}

\newcommand{\Papprox}{\tilde{\mathcal{P}}}

\newcommand*{\eg}{e.g.\@\xspace}
\newcommand*{\ie}{i.e.\@\xspace}

\newcommand{\egreedy}{\ensuremath{\epsilon\text{-greedy}}\xspace}
\newcommand{\eshotgun}{\ensuremath{\epsilon\text{-shotgun}}\xspace}
\newcommand{\eSRS}{\ensuremath{\epsilon\text{S-RS}}\xspace}
\newcommand{\eSPF}{\ensuremath{\epsilon\text{S-PF}}\xspace}
\newcommand{\eExploit}{\ensuremath{\epsilon\text{S-0}}\xspace}

\newcommand{\pushfour}{\ensuremath{\textsc{push4}}\xspace}
\newcommand{\pusheight}{\ensuremath{\textsc{push8}}\xspace}

\copyrightyear{2020} 
\acmYear{2020} 
\setcopyright{acmlicensed}
\acmConference[GECCO '20]{Genetic and Evolutionary Computation Conference}{July 8--12, 2020}{Cancún, Mexico}
\acmBooktitle{Genetic and Evolutionary Computation Conference (GECCO '20), July 8--12, 2020, Cancún, Mexico}
\acmPrice{15.00}
\acmDOI{10.1145/3377930.3390154}
\acmISBN{978-1-4503-7128-5/20/07}

\begin{document}
\title{\eshotgun: \egreedy Batch Bayesian Optimisation}

\author{George {De Ath}}
\email{g.de.ath@exeter.ac.uk}
\orcid{0000-0003-4909-0257}
\affiliation{%
  \department{Department of Computer Science}
  \institution{University of Exeter}
  \city{Exeter}
  \country{United Kingdom}
}

\author{Richard M. Everson}
\email{r.m.everson@exeter.ac.uk}
\orcid{0000-0002-3964-1150}
\affiliation{%
  \department{Department of Computer Science}
  \institution{University of Exeter}
  \city{Exeter}
  \country{United Kingdom}
}

\author{Jonathan E. Fieldsend}
\email{j.e.fieldsend@exeter.ac.uk}
\orcid{0000-0002-0683-2583}
\affiliation{%
  \department{Department of Computer Science}
  \institution{University of Exeter}
  \city{Exeter}
  \country{United Kingdom}
}

\author{Alma A. M. Rahat}
\email{a.a.m.rahat@swansea.ac.uk}
\orcid{0000-0002-5023-1371}
\affiliation{%
  \department{Department of Computer Science}
  \institution{Swansea University}
  \city{Swansea}
  \country{United Kingdom}
}

\ccsdesc[500]{Theory of computation~Gaussian processes}
\ccsdesc[500]{Theory of computation~Mathematical optimization}
\ccsdesc[500]{Computing methodologies~Modeling and simulation}
\ccsdesc[500]{Computing methodologies~Optimization algorithms}

\keywords{Bayesian optimisation, Batch, Parallel, Exploitation, 
         \egreedy, Infill criteria, Acquisition function}

\begin{abstract}
  Bayesian optimisation is a popular surrogate model-based approach for
  optimising expensive black-box functions. Given a surrogate model, the
  next location to expensively evaluate is chosen via maximisation of a
  cheap-to-query acquisition function. We present an \egreedy procedure for
  Bayesian optimisation in batch settings in which the black-box function
  can be evaluated multiple times in parallel. Our \eshotgun algorithm
  leverages the model's prediction, uncertainty, and the approximated rate
  of change of the landscape to determine the spread of batch solutions to
  be distributed around a putative location. The initial target location is
  selected either in an exploitative fashion on the mean prediction, or --
  with probability $\epsilon$ -- from elsewhere in the design space. This
  results in locations that are more densely sampled in regions where the
  function is changing rapidly and in locations predicted to be good (\ie
  close to predicted optima), with more scattered samples in regions where
  the function is flatter and/or of poorer quality. We empirically evaluate
  the \eshotgun methods on a range of synthetic functions and two
  real-world problems, finding that they perform at least as well as
  state-of-the-art batch methods and in many cases exceed their
  performance.
\end{abstract}

\maketitle

\section{Introduction}
\label{sec:introduction}
Global optimisation of non-convex and black-box functions is a common task in many
real-world problems. These include hyperparameter tuning of machine learning
algorithms \citep{snoek:practical}, 
drug discovery \citep{hernandez-lobato:chemical},
analog circuit design \citep{lyu:circuit},
mechanical engineering design
\citep{chugh:airintake, daniels:diffuser, rahat:coalboiler}
and general algorithm configuration \citep{hutter:algconf}. Bayesian 
optimisation (BO) has become a popular approach for optimising expensive, 
black-box functions that have no closed-form expression or derivative 
information \citep{snoek:practical, shahriari:ego}. It employs a  probabilistic surrogate model of a function using available
function evaluations. The location at which the function is next expensively
evaluated is chosen as the location that maximises an acquisition function
(or infill criterion) that balances exploration and exploitation.

In real-world problems it is often possible to run multiple experiments in
parallel by using modern hardware capabilities to expensively evaluate several
locations at once. When optimising machine learning algorithms, for example, 
multiple model configurations can be evaluated in parallel across many 
processor cores on one or multiple machines \citep{chan:alphago, kandasamy:ts}.
Consequently, this has led to the development of batch (or parallel) BO 
algorithms, which use acquisition functions to select $q$ locations to be 
evaluated at each iteration. Clearly, a strictly serial evaluation makes the
best overall use of the available CPU time because each new
location to be evaluated is selected with the maximum available
information.  Parallel evaluation, however, holds the promise of
substantially reducing the wall-clock time to locate the optimum.  

The selection of a good set of locations to evaluate at each batch iteration is
a non-trivial problem. In sequential BO, techniques which favour
greedy exploitation of the surrogate model have been shown to be preferable to the
more traditional acquisition functions \citep{rehbach:ei_pv,death:egreedy}. \citet{death:egreedy}, for example, show that using an \egreedy 
strategy of exploiting the surrogate model the majority of the time and, with
probability $\epsilon$ (where $\epsilon \approx 0.1$), randomly selecting a 
location to explore yields superior optimisation performance on a variety of
synthetic and real-world problems. Consequently, in this work we
investigate \egreedy methods in the batch BO setting.

We present \textit{\eshotgun}, a novel approach to batch BO,  which uses an \egreedy
strategy for selecting the first location $\xnext_1$
in a batch, and then samples
the remaining $q-1$ points from a normal distribution centred on $\xnext_1$, with
a scale parameter determined by the surrogate model's posterior mean and
variance at $\xnext_1$ and the magnitude of the gradient in the vicinity of
$\xnext_1$. This embodies maximum exploitation of the surrogate model the
majority of the time by virtue of the choice of $\xnext_1$.  The remaining
$q-1$ locations may be exploratory or exploitative 
depending on the characteristics of the local landscape. Larger regions of
decision space
will be sampled when $\xnext_1$ is surrounded by a relatively flat landscape,
while denser sampling will occur where $\xnext_1$ is in a locally steeper 
region, such as the landscape around a local (or global) optimum.

Our contributions can be summarised as follows:
\begin{itemize}
\item We present \eshotgun, a new batch Bayesian optimisation approach based on
      the \egreedy strategy of exploiting the surrogate model.
\item We empirically compare a range of state-of-the-art batch Bayesian optimisers across a variety of 
	  synthetic test problems and two real-world applications
\item We empirically show that the \eshotgun approaches are equal to or better
	  than several state-of-the-art batch BO methods on a wide range of problems.
\end{itemize}

We begin in Section~\ref{sec:review} by briefly reviewing Bayesian optimisation
along with Gaussian processes (the surrogate model generally used in BO) 
and common acquisition functions. Batch BO and the algorithm archetypes used 
for selecting the batch locations are then reviewed in
Section~\ref{sec:review:batchbo}, which leads to the proposed \eshotgun
approach in Section~\ref{sec:eshotgun}. Empirical evaluation on well-known test problems and two real-world applications are
presented in Section~\ref{sec:exps}. We finish with concluding remarks in
Section~\ref{sec:conc}.

\section{Bayesian Optimisation}
\label{sec:review}
Our goal is to minimise a black-box function $f : \mX \mapsto \Real$, defined
on a compact domain $\mX  \subset \Real^d$. The function itself is unknown, but
we have access to the results of its evaluations $f(\bx)$ at any location
$\bx \in \mX$. We are particularly interested in cases where the evaluations
are expensive, either in terms of time or money or both, and we seek to
minimise $f$ in either as few evaluations as possible to incur as little cost
as possible or for a fixed budget.

\subsection{Sequential Bayesian Optimisation}
\label{sec:review:bo}
Bayesian Optimisation (BO), also known as Efficient Global Optimisation, is a
global search strategy that sequentially samples design space at locations that
are likely contain the global optimum, taking into account the predictions of
the surrogate model and their associated uncertainty \citep{jones:ego}. It 
starts by generating $M$ initial sample locations $ \{\bx_i\}_{i=1}^M$ with a space 
filling algorithm, typically Latin hypercube sampling \citep{mckay:lhs}, and 
expensively evaluates them with the function, $f_i = f(\bx_i)$. This collected
set of observations forms the dataset with which the surrogate model is 
initially trained. Following model training, and at each iteration of BO, the 
next location for expensive evaluation is selected according to an acquisition 
function (or infill criterion). These usually combine the surrogate model's 
prediction and prediction uncertainty of  the design space to balance the exploitation of
promising solutions (those with good predicted values) and those solutions with
high uncertainty. The location $\xnext$ maximising this criterion is used as 
the next point to be expensively evaluated. The dataset is augmented with
$\xnext$ and $f(\xnext)$ and the process is repeated until the budget is
exhausted. The value of the global minimum $\fmin$ is estimated to be the best
function evaluation seen during the optimisation run, \ie
$\fstar = \min_i \{ f_i \}$.

\subsubsection{Gaussian Processes}
\label{sec:review:gp}
Gaussian processes (GP) are a popular and versatile choice of surrogate model
for $f(\bx)$, due to their strengths in function approximation and uncertainty
quantification
\citep{rasmussen:gpml}. A GP is a collection of random variables, and any 
finite number of these are jointly Gaussian distributed. A GP
prior over $f$ can be defined as $\mGP( m(\bx), \kappa(\bx, \bx' \given \btheta))$ where
$m(\bx)$ is the mean function, $\kappa(\cdot, \cdot)$ is the kernel function
(also known as a covariance function) and $\btheta$ are the hyperparameters of
the kernel. Given data consisting of $f(\bx)$ evaluated at $M$ sampled
locations $\Data = \{ (\bx_i, f_i \triangleq f(\bx_i)) \}_{i=1}^{M}$, the 
posterior estimate of $f$ at location $\bx$ is a Gaussian distribution:
\begin{equation}
p ( f(\bx) \given \bx, \Data, \btheta) = \Normal ( \mu(\bx), \sigma^2(\bx) )
\end{equation}
with mean and variance
\begin{align}
\mu(\bx \given \Data, \btheta)
  &= \bkappa( \bx, X ) K^{-1} \mathbf{f} \\
\sigma^2(\bx \given \Data, \btheta)
  &= \kappa(\bx, \bx) - \bkappa( \bx, X)^\top K^{-1} \kappa(X, \bx),
\end{align}
where $X \in \Real^{M \times d}$ is matrix of input locations in each row and
$\mathbf{f} \in \Real^{M}$ is the corresponding vector of true function evaluations
$\{f_1, f_2, \dots, f_M\}$. The matrix $K \in \Real^{M \times M}$
contains the kernel evaluated at each pair of observations, and
$\bkappa( \bx, X )$ is the  $M$-dimensional vector  whose elements are $[\bkappa(\bx, X)]_i
= \kappa(\bx, \bx_i)$. Kernel hyperparameters $\btheta$ are learnt via
maximising the log likelihood:
\begin{equation}
\log p (\Data \given \btheta) = - \frac{1}{2} \log \lVert K \rVert 
                               - \frac{1}{2} \mathbf{f}^\top K^{-1} \mathbf{f}
                               - \frac{M}{2} \log (2\pi).
\label{eqn:kernel_log_likelihood}
\end{equation}
For notational simplicity, we drop explicit dependencies on the data $\Data$
and kernel hyperparamters $\btheta$ from now on.

\subsubsection{Acquisition Functions}
\label{sec:review:acqfuncs}
An acquisition function $\alpha(\bx)$ is used to measure the anticipated quality of
expensively evaluating $f$ at any given location $\bx$: the location that
maximises the acquisition function is chosen as the next location for
expensive evaluation. While this strategy may appear merely to transfer the
problem of optimising $f(\bx)$ to a maximisation of $\alpha(\bx)$,
the acquisition function is cheap to evaluate so the location of its global
optimum can be cheaply estimated using an evolutionary algorithm. 

Acquisition functions attempt, either implicitly or explicitly, to balance
the trade-off
between maximally exploiting the surrogate model, \ie selecting a location
with the best predicted value, and maximally exploring the model, \ie selecting
the location with the most uncertainty. Perhaps the two most widespread
acquisition functions, are Expected Improvement (EI) \citep{jones:ego} and
Upper Confidence Bound (UCB) \citep{srinivas:ucb}. EI measures the positive
predicted improvement over the best solution observed so far and UCB is a
weighted sum of the surrogate model's mean prediction $\mu(\bx)$ and 
uncertainty $\sigma^2(\bx)$. These were both shown \citep{death:egreedy} to be
monotonic with respect to increases in both $\mu(\bx)$ and $\sigma^2(\bx)$ and
that the solutions that maximise them both belong to the Pareto set of
locations which maximally
trade-off   exploitation (minimising $\mu(\bx)$) and  exploration 
(maximising  $\sigma^2(\bx)$).

Recently, \egreedy approaches have been successfully used as
acquisition functions \citep{death:egreedy}. These select a maximally
exploitative solution, $\xnext = \argmin_x \, \mu(\bx)$ with probability 
$1-\epsilon$ and select a random solution with probability $\epsilon$.
\citet{death:egreedy} present two methods for selecting the random solution,
either uniformly from $\mX$ or from the approximate Pareto set of solutions
of the surrogate model's mean prediction and variance. They showed that \egreedy
approaches are particularly effective on higher dimensional problems, and that performing
pure exploitation (\ie $\epsilon = 0$) is competitive with the best-performing
methods. This result was recently confirmed by \citet{rehbach:ei_pv}, who
empirically show that solely using the surrogate model's predicted value 
performs better than EI on most problems with a dimensionality of 5 or more.

\subsection{Batch Bayesian Optimisation}
\label{sec:review:batchbo}
In batch Bayesian optimisation (BBO) the goal is to select a batch
$\mX' = \{\xnext_1, \dots, \xnext_q\}$ of $q$ promising locations to expensively
evaluate in parallel. 
 One of the earliest BBO approaches, the qEI method of 
\citet{ginsbourger:qei}, generalised the sequential EI acquisition function to 
a batch setting in which all $q$ batch locations are jointly estimated.
However, it is not analytically tractable to compute qEI, even for
small batch sizes \citep{gonzalez:lp}.  Although a fast approximation to qEI does
exist \citep{chevalier:fast-qei}, it is not faster than naive Monte Carlo
approximations for larger batch sizes. More recently, \citet{wang:moe_qei}
proposed a more efficient algorithm to estimate the gradient of qEI,
but the approach still results in having to optimise in a $d \times q$ dimensional
space for each set of batch locations. Two other methods that jointly optimise 
the batch of locations, the parallel predictive entropy search
\citep{shah:ppes} and the parallel knowledge gradient method \citep{jian:pkgm},
have also been shown to scale poorly as batch size increases
\citep{daxberger:distbbo}.

Consequently, iteratively selecting the batch sample locations has become the 
prevailing
methodology. One such strategy is to attempt to ensure that different locations are
selected for the batch by, for each of the $q$ locations, sampling a realisation from  the
surrogate model posterior (Thompson Sampling) and minimising it \citep{kandasamy:ts}. 
However, this relies on there being sufficient uncertainty in the model to 
allow for the realisations to have different minima 
\citep{depalma:tsacuisition}. 
\citet{depalma:tsacuisition} proposed sampling from a distribution of
acquisition functions, or rather from the distribution of hyperparameters that
control the acquisition function's behaviour, such as the trade-off between
exploration and exploitation in UCB.

Instead of relying on the stochasticity of either the surrogate model or 
acquisition function hyperparameters, another group of methods penalise the 
regions from which a batch point has already been selected; thus they are less 
likely (or unable to) select from nearby locations. A well-known heuristic to
achieve this is to \textit{hallucinate} the results of pending evaluations
\citep{azimi:hallu, ginsbourger:kb, desautels:bucb}. In this set of methods,
the first batch location is selected by optimising an acquisition function and
then subsequent locations are chosen by incorporating the predicted outcome of
the already-selected batch locations into the surrogate model and optimising 
the acquisition function over the new model. The popular
Kriging Believer method \citep{ginsbourger:kb} uses the surrogate's mean prediction
as the hallucinated value, which reduces the model's posterior uncertainty
to zero at the hallucinated locations without affecting the posterior mean.

An alternative to penalising the surrogate model via hallucination is to 
penalise an acquisition function in a region around the selected batch points
\citep{gonzalez:lp, alvi:asynclp}. In these methods, the first point $\bx'_1$ in a batch
is selected via maximisation of a sequential acquisition function 
$\alpha(\bx)$, \eg EI; the subsequent $q-1$ locations are chosen by 
iteratively maximising a penalised version of the sequential acquisition
function:
\begin{equation}
  \bx'_i = \underset{\bx \in \mX}{\argmax} 
  \left\lbrace
    \alpha(\bx) \prod_{j=1}^{i-1} \varphi(\bx \given \bx'_j)
  \right\rbrace,
  \quad i = 2, \ldots, q
\end{equation}
where $\varphi(\bx \given \bx_j)$ are local penalisers centred at $\bx_j$.
These penalise a region around $\bx_j$ with decreasing penalisation as the
distance from $\bx_j$ increases; for example \citep{gonzalez:lp} use a
squared exponential function.  The length scale over which the penalisation
is significant is set by 
\begin{equation}
r_j = \frac{| \mu(\bx'_j) - \fmin |}{L} + \gamma \frac{\sigma^2(\bx'_j)}{L},
\label{eqn:ball_radius}
\end{equation}
where $\fmin$ is equal to the global minimum of the function, $L$ is a valid
Lipschitz constant expressing how rapidly $f$ can change with $\bx$, and $\gamma \geq 0$ weights the importance of the
uncertainty about $\bx_j$. 

In practise, the true value of $\fmin$ is unknown and therefore the
best seen value so far, $\fstar = \min_i \{ f_i \}_{i=1}^M$, is used in lieu. It can be shown 
\citep{gonzalez:lp} that 
$L_\nabla = \max_{\bx \in \mX} \lVert \nabla f(\bx) \rVert$ is a
valid Lipschitz constant and \citet{gonzalez:lp} approximate this using the 
surrogate model's mean prediction: 
$\tilde{L} = \max_{\bx \in \mX} \lVert \nabla \mu(\bx) \rVert$, resulting in a
global estimate of the largest gradient in the model that is fixed for all
selected batch locations. \citet{alvi:asynclp} argue that this under-penalises
flatter regions of space in which the estimated gradient of the function is
much smaller, and therefore they calculate a different value of $\tilde{L}$ for
each selected batch location, estimating it within a length-scale of each 
location. \citet{gonzalez:lp} set $\gamma = 0$ and focus only on the
difference between the predicted value of $\bx_j$ and the global optimum,
whereas \citet{alvi:asynclp} let $\gamma = 1$ to also include prediction
uncertainty. This penalty  shrinks as the 
predicted value of $\bx_j$ approaches the global minimum and also as the 
largest local (or global) gradient of the model increases.

Motivated by the success of the sequential exploitative and 
\egreedy approaches \citep{death:egreedy, rehbach:ei_pv}, we invert the local
penalisation strategy and, instead, present a method that samples from 
\textit{within} the region that would usually be penalised \eqref{eqn:ball_radius}. We 
empirically show that this approach out-performs recent BBO methods on a range
of synthetic functions and two real-world problems.

\section{$\epsilon$-Shotgun BBO}
\label{sec:eshotgun}
Motivated by the recent success of \egreedy methods, we extend the two 
sequential methods of \citet{death:egreedy} to the batch setting. We use an
 \egreedy acquisition function to generate the first
batch location $\xnext_1$ and then sample the remaining locations from a normal distribution
centred on $\xnext_1$, with a standard deviation given by:
\begin{equation}
r = \frac{| \mu(\bx'_1) - \fmin |}{L} + \gamma \frac{\sigma^2(\bx'_1)}{L}.
\label{eqn:sample-radius}
\end{equation}
Sampling in this manner creates a scattered set of batch points around $\xnext_1$,
akin to a shotgun blast, whose approximate spread is determined by the amount
of model uncertainty of $\xnext_1$, its predicted value relative to the 
best seen function evaluation, and by the steepest gradient within its
vicinity.

\begin{algorithm}[t]
\caption{\eshotgun query point selection for BBO}
\label{alg:eshotgun}
 \begin{algorithmic}[]
 	\State \textbf{Inputs:}
 	\State {\setlength{\tabcolsep}{2pt}%
 	        \begin{tabular}{c p{2pt} l}
 			$q$ &:& Batch size \\
 			$\epsilon$ &:& Proportion of the time to explore \\
 			$l$ &:& Kernel length scale
 			\end{tabular}
 	       }%
         \end{algorithmic}
\medskip

\begin{algorithmic}[1]
	\If{ $\mathtt{rand()} < \epsilon$ }
		\If {Using Pareto front selection}
		\Comment{\eSPF}
			\State $\Papprox \gets \mathtt{MOOptimise}_{\bx\in\mX}(\mu(\bx), \sigma^2(\bx))$
			\label{alg:eS:moo_optimise}
	        \State{$\xnext_1 \gets \mathtt{randomChoice}(\Papprox)$}
			\label{alg:eS:choice_PF}
		\Else
		\Comment{\eSRS}
	    	\State{$\xnext_1 \gets \mathtt{randomChoice}(\mX)$}
	    	\label{alg:eS:choice_random}
		\EndIf
    \Else
		\State $\xnext_1 \gets \underset{\bx \in \mX}{\argmin} \, \mu(\bx)$
			\label{alg:eS:exploit_model}
	\EndIf
	
	\State $\tilde{L} = \underset{\bx \in [\xnext_1 - l, \; \xnext_1 + l]^d}{\max} 
	                    \, \left\lVert \mu_{\nabla} (\bx) \right\rVert$
	\Comment{Largest gradient; centred on $\xnext_1$}
	\label{alg:es:grad_est}
	
	\State $\fstar = \min \left\lbrace f_i \right\rbrace_{i=1}^M$
	\Comment{Best seen function value}

	\State $r = \frac{|\mu(\xnext_1) - \fstar|}{\tilde{L}}
	            + \gamma \frac{\sigma(\xnext_1)}{\tilde{L}}$
	\State $\mX' = \left\lbrace \xnext_1 \right\rbrace$
	
	\medskip
	
	\While{ $\lvert \mX' \rvert < q$ }
	\label{alg:es:sample_points_start}
		\State $\bx' \sim \Normal (\xnext_1, r^2\bI)$
		\If { $\bx' \in \mX$ }
			\State $\mX' \gets \mX' \cup \left\lbrace \bx' \right\rbrace$
		\EndIf
	\label{alg:es:sample_points_end}
	\EndWhile
	
	\State \Return $\mX'$
\end{algorithmic}
\end{algorithm}
We describe two alternative strategies employing this idea, which are summarised
in Algorithm~\ref{alg:eshotgun}.
The first method, which we call \textit{\eshotgun with Pareto front selection}
(\eSPF), selects with probability
$1 - \epsilon$  the location $\xnext_1$ with the most promising mean
prediction from the surrogate model (line~\ref{alg:eS:exploit_model}). In the
remaining cases it selects a random element from
the approximate Pareto set $\Papprox$, which is found using an evolutionary
multi-objective optimiser  
(lines~\ref{alg:eS:moo_optimise} and \ref{alg:eS:choice_PF}).

Following the selection of the $\xnext_1$, \eSPF samples the remaining $q-1$
locations from a normal distribution $\Normal (\xnext_1, r^2\bI)$ centred on 
$\xnext_1$ with a standard deviation equal to the radius 
\eqref{eqn:sample-radius} of penalisation used in \citep{alvi:asynclp}
(lines \ref{alg:es:grad_est} to \ref{alg:es:sample_points_end}). We 
conservatively estimate the global optimum $\fmin$ to be the best function evaluation 
seen so far, \ie $\fstar = \min_i \{ f_i \}$. The localised Lipschitz
constant, similarly to \citet{alvi:asynclp}, is estimated to be
$\tilde{L} = \max_{\bx \in \mathcal{H}} \lVert \nabla \mu(\bx) \rVert$,
where $\mathcal{H}$ is a hypercube, centred on $\xnext_1$ with side lengths of 
twice the length-scale of the surrogate model's kernel. This allows for the 
local gradient to influence the size of the sampling region. If the local
gradient is small then it is beneficial to sample over a wide region to
learn more about the 
structure of $f$. Conversely, a steeper local
gradient would indicate that the modelled function is changing rapidly and 
therefore sampling more densely (due to a larger $\tilde{L}$) is required
to accurately model $f$ and guide the search.

\begin{figure}[t]
\includegraphics[width=\columnwidth, clip, trim={0 0 0 0}]{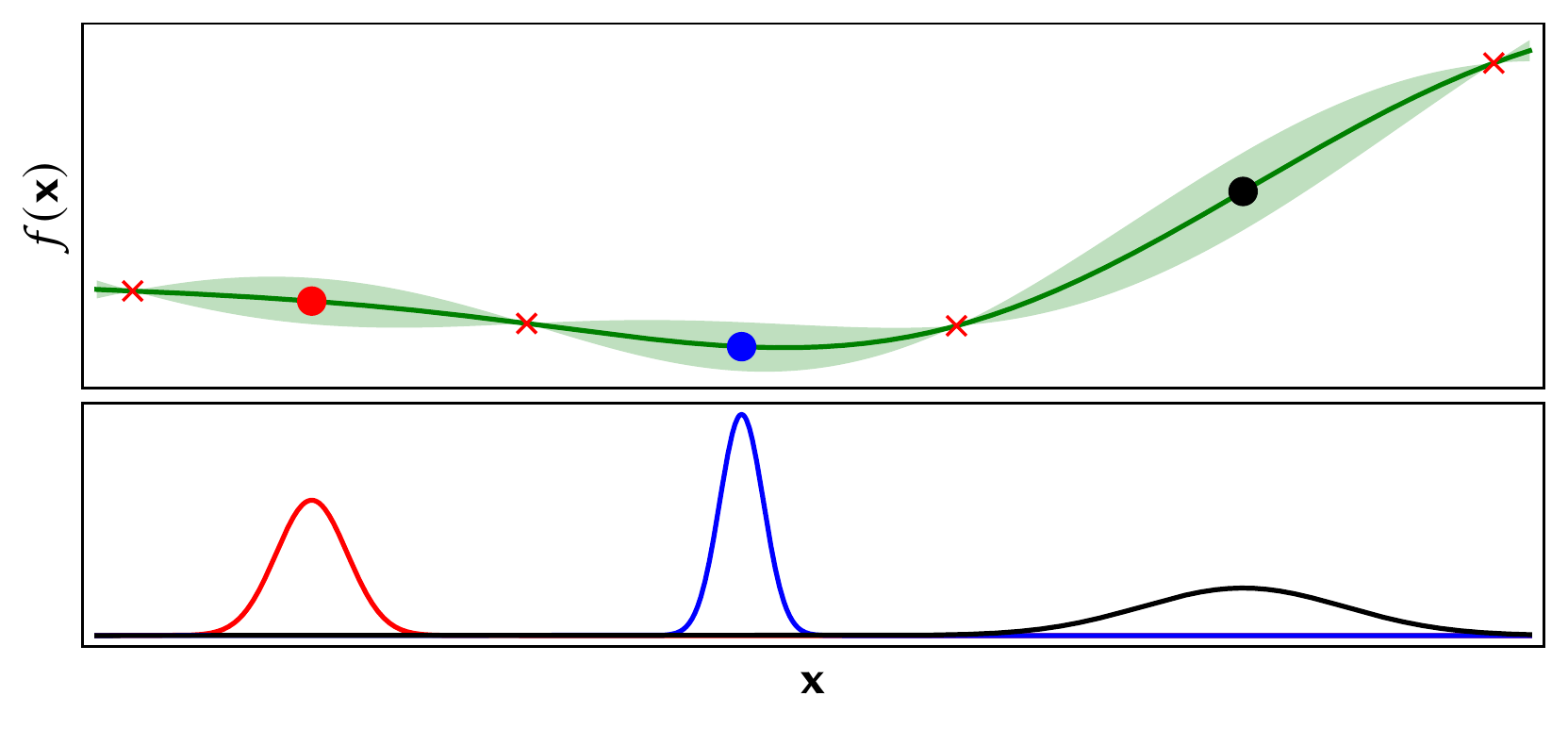}%
\caption{\eshotgun selection example. The upper panel shows the predicted mean and
uncertainty (green) of an unknown function sampled at four locations 
(red crosses). The lower panel shows the pdfs of $\Normal (\xnext_1, r^2)$, 
centred on the three correspondingly coloured locations of $\xnext_1$ in the 
upper figure.}
\label{fig:eshotgun_radius_example}
\end{figure}
Figure~\ref{fig:eshotgun_radius_example} shows three example locations of $\xnext_1$ for a surrogate model and their corresponding probability
density functions from which samples would be drawn.
The blue circle is located at the minimum of the modelled 
function, and its corresponding sampling radius is relatively small because
$| \mu(\xnext_1) - \fstar |$ is small, the local gradient is small, and the
predicted uncertainty is relatively large, resulting in a fairly sharp
distribution to sample from. The red circle corresponds to a location with a
sampling radius that is larger than the previous point, because, while the 
 difference between the  modelled function and  $\fstar$ and the predicted uncertainty is similar to the blue location, the local
gradient is smaller. Lastly, the black location has a similar model 
uncertainty to the red location, but has a much larger $|\mu(\xnext_1) -
\fstar|$  resulting in a much wider pdf, even when taking into
account the larger local gradient.

The second strategy, \textit{\eshotgun with random selection} (\eSRS), is
identical to \eSPF except that, with probability $\epsilon$, it selects
$\xnext_1$ at a random location in the entire feasible space
(line~\ref{alg:eS:choice_random}), instead of a location from the
approximate Pareto set. It might be expected that \eSPF would outperform
\eSRS because it selects locations that are more informative to the
optimisation process, because they  are non-dominated with
respect to their predicted value and uncertainty. However,
since \citet{death:egreedy} show that sequential \egreedy methods based on
selection from the Pareto set have only marginally better performance than
purely random selection,  we  include \eSRS to assess whether this
is true in the batch setting.

\section{Experimental Evaluation}
\label{sec:exps}
We investigate the performance of the two proposed \eshotgun methods,
\eSPF and \eSRS on ten well-known benchmark functions with 
a differing dimensionality and two real-world applications in the form of an
active learning problem for robot pushing and pipe shape optimisation. Full
results of all experimental evaluations are available in the supplementary
material.

Following the reported success of purely exploitative methods
\citep{death:egreedy, rehbach:ei_pv}, we also compare \eSPF and \eSRS to the purely exploitative \eshotgun method without any
random point selection (\ie $\epsilon = 0$), which we denote 
\eExploit. We also compare 
the batch \eshotgun methods to five %
BBO methods representative of different styles of batch optimisation as
discussed in Section~\ref{sec:review:batchbo}:   Two acquisition function-based 
penalisation methods: the popular Local Penalisation (LP) method 
\citep{gonzalez:lp}, which uses soft penalisation
($\varphi(\bx_j \given  \bx_j) > 0$), and the more recent 
PLAyBOOK \citep{alvi:asynclp} method that uses hard local penalisation 
($\varphi(\bx_j \given \bx_j) = 0$). Kriging Believer (KB),
which penalises by hallucinating the already-selected batch points
\citep{ginsbourger:kb}; and the Thompson sampling (TS) method of
\citet{kandasamy:ts}, which minimises a realisation of the modelled function
from the surrogate. Lastly, we include qEI \citep{ginsbourger:qei}, which jointly 
estimates the location of the $q$ batch members. All methods were implemented
in Python using the same 
packages\footnote{Implementation available: \url{https://github.com/georgedeath/eshotgun}},
apart from the local penalisers of LP and PLAyBOOK which used the
PLAyBOOK implementation.\footnote{\url{https://github.com/a5a/asynchronous-BO}}

A zero-mean Gaussian process surrogate model with an isotropic Mat{\'e}rn $5/2$
kernel was used in all the  experiments. The kernel was selected due 
to its widespread usage and recommended use for modelling realistic functions
\citep{snoek:practical}. The models were initially trained on $2d$ observations
generated by maximin Latin hypercube sampling \citep{mckay:lhs}, with each
optimisation run repeated $51$ times with different initialisations.
The same sets of initial batch locations were common across all methods
to enable statistical comparison. At each iteration, before batch point 
selection, the hyperparameters of the GP were optimised by maximising the log
likelihood \eqref{eqn:kernel_log_likelihood} with L-BFGS-B \citep{byrd:lbfgs}
using 10 restarts \citep{gpy}.

The LP, PLAyBOOK and KB methods all used the EI acquisition function. For each
location selected in LP and PLAyBOOK, we followed the authors' guidelines \citep{alvi:asynclp} and
uniformly sampled the acquisition function at $3000$ locations, selecting the
best location after locally optimising (with L-BFGS-B) the best 5. For
the other methods, a maximum budget of $10000d$ acquisition function 
evaluations was used in conjunction with L-BFGS-B for functions with $d=1$ and
for $d \geq 2$ we used CMA-ES using the standard bi-population strategy \citep{hansen:bipop} and
(up to) 9 restarts. The approximate Pareto set $\Papprox$ of non-dominated 
locations (in terms of $\mu(\bx)$ and $\sigma^2(\bx)$) in \eSPF was found using
NSGA-II \citep{deb:nsga2} with a $100d$ population size, $d^{-1}$ %
mutation rate, $0.8$ crossover rate, and crossover and mutation distribution
indices of $\eta_c = \eta_m = 20$. For 
both \eSRS and \eSPF we took $\epsilon = 0.1$.

\subsection{Synthetic Experiments}
\label{sec:exps:synthetic}
\begin{table}[t]
\centering
\begin{tabular}[t]{lr c lr}
\addlinespace[-\aboverulesep]\cmidrule[\heavyrulewidth]{1-2}\cmidrule[\heavyrulewidth]{4-5}
	\bfseries Name                                & $d$ && \bfseries Name                & $d$  \\
\cmidrule[\lightrulewidth]{1-2}\cmidrule[\lightrulewidth]{4-5}
	WangFreitas \citep{wangfreitas:theoreticalBO} & 1   && logSixHumpCamel$^\dagger$     & 2    \\
	Branin$^\dagger$                              & 2   && modHartman6$^\dagger$         & 6    \\
	BraninForrester \citep{forrester:engdesign}   & 2   && logGSobol \citep{gonzalez:lp} & 10   \\
	Cosines \citep{gonzalez:glasses}              & 2   && logRosenbrock$^\dagger$       & 10   \\
	logGoldsteinPrice$^\dagger$                   & 2   && logStyblinkskiTang$^\dagger$  & 10   \\
\cmidrule[\heavyrulewidth]{1-2}\cmidrule[\heavyrulewidth]{4-5}\addlinespace[-\belowrulesep]
\end{tabular}
\caption{Synthetic functions used and their dimensionality $d$. Formulae can be
found as cited or at \url{http://www.sfu.ca/~ssurjano/optimization.html} for
those labelled with $\dagger$.}
\label{tbl:function_details}
\end{table}
The methods were evaluated on the 10 synthetic benchmark functions in 
Table~\ref{tbl:function_details} with batch sizes $q \in \{2,5,10,20\}$ and a
fixed budget of $200$ function evaluations. Table~\ref{tbl:synthetic_results}
shows, for a batch size of $q=10$, the median difference (over $51$
repeated experiments) between the estimated
optimum $\fstar$ and true optimum, as well
as the median absolute deviation from the median (MAD), a robust measure of
dispersion. The method with the minimum median $\fstar$ for each function is 
highlighted in dark grey, and those that are statistically equivalent to the best
method according to a one-sided, paired Wilcoxon signed-rank test
\citep{knowles:testing} with Holm-Bonferroni correction \citep{holm:test}
($p\geq0.05$), are shown in light grey. Note that tabulated results for all batch
sizes are available in the supplementary material.

\begin{figure*}[t]
\includegraphics[width=0.5\textwidth, clip, trim={0 0 0 0}]{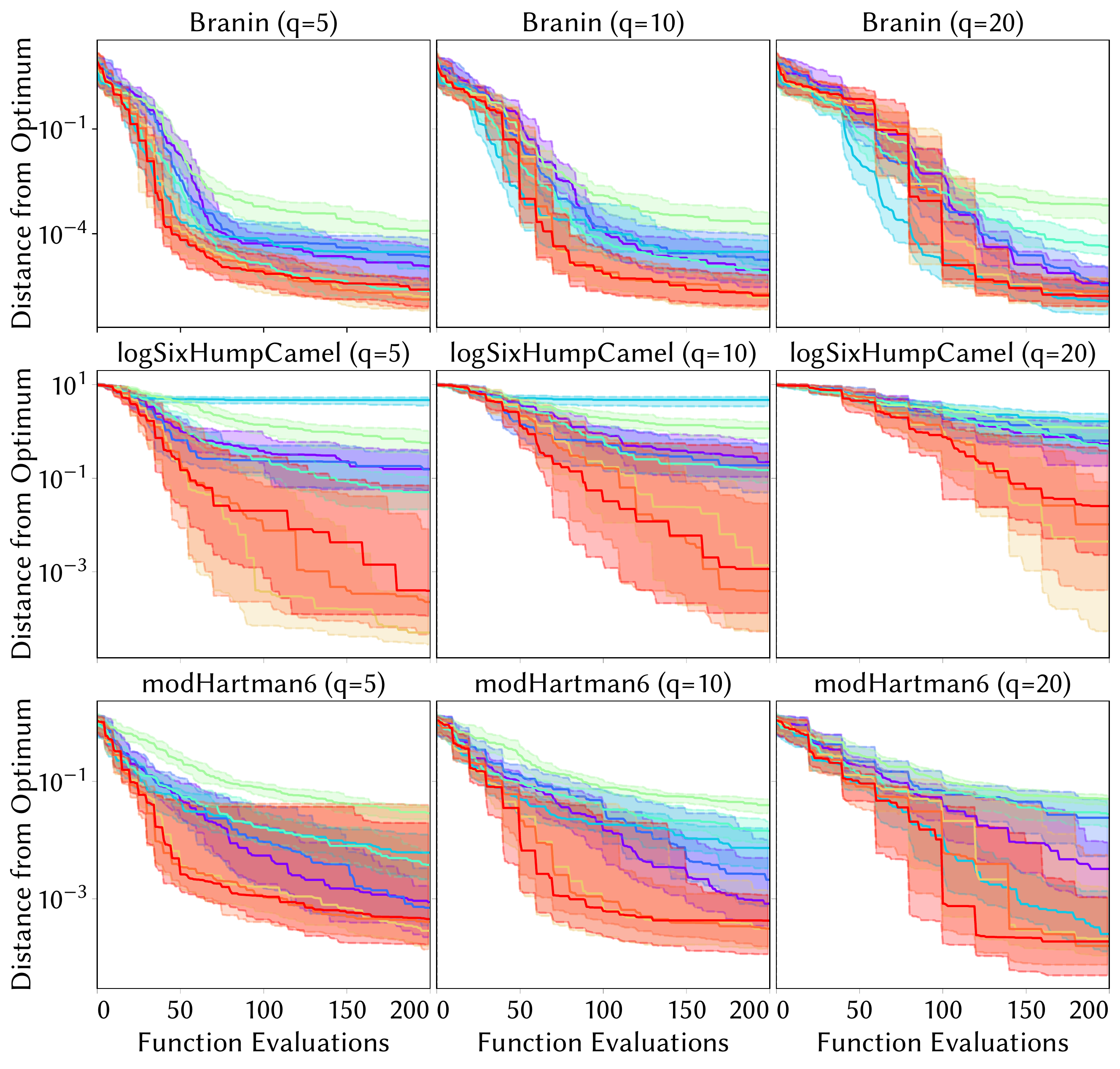}%
\includegraphics[width=0.5\textwidth, clip, trim={0 0 0 0}]{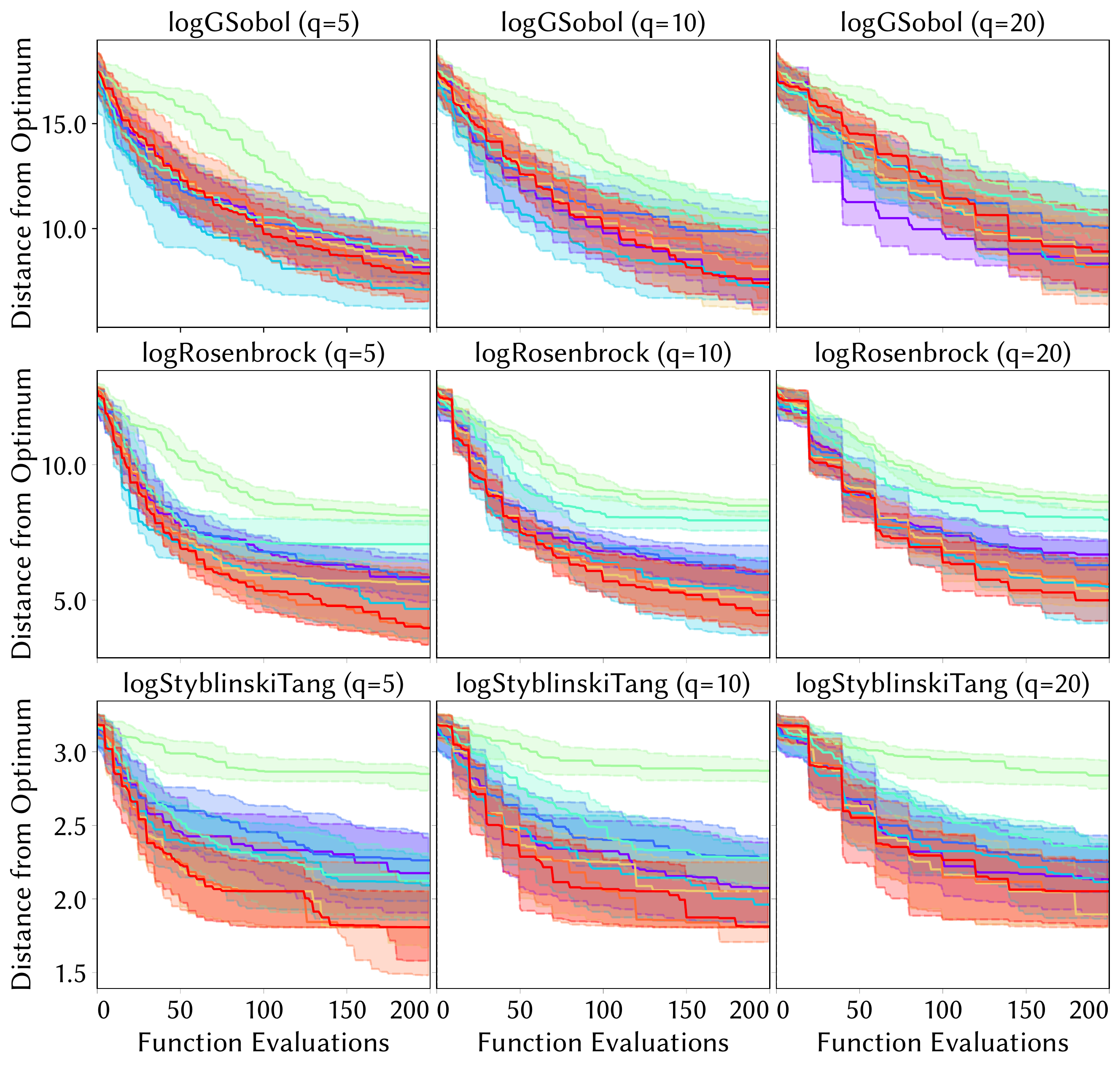}\\
\includegraphics[width=0.8\textwidth, clip, trim={10 10 10 10}]{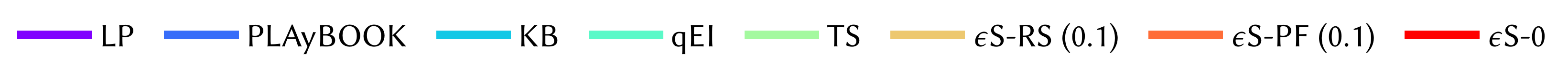}%
\caption{Illustrative convergence plots for six benchmark problems and three
batch sizes $q \in \{5,10,20\}$. Each plot shows the median difference between
the best function value seen $\fstar$ and the true optimum $\fmin$, with shading representing
the interquartile range across the 51 runs.}
\label{fig:conv_plots}
\end{figure*}

\begin{table*}[t]
\setlength{\tabcolsep}{2pt}
\sisetup{table-format=1.2e-1,table-number-alignment=center}
\resizebox{1\textwidth}{!}{%
  \begin{tabular}{l | SS| SS| SS| SS| SS}
    \toprule
    \bfseries Method
    & \multicolumn{2}{c|}{\bfseries WangFreitas (1)} 
    & \multicolumn{2}{c|}{\bfseries BraninForrester (2)} 
    & \multicolumn{2}{c|}{\bfseries Branin (2)} 
    & \multicolumn{2}{c|}{\bfseries Cosines (2)} 
    & \multicolumn{2}{c}{\bfseries logGoldsteinPrice (2)} \\ 
    & \multicolumn{1}{c}{Median} & \multicolumn{1}{c|}{MAD}
    & \multicolumn{1}{c}{Median} & \multicolumn{1}{c|}{MAD}
    & \multicolumn{1}{c}{Median} & \multicolumn{1}{c|}{MAD}
    & \multicolumn{1}{c}{Median} & \multicolumn{1}{c|}{MAD}
    & \multicolumn{1}{c}{Median} & \multicolumn{1}{c}{MAD}  \\ \midrule
    LP & 2.00e+00 & 3.08e-09 & 2.61e-05 & 3.86e-05 & 9.25e-06 & 1.23e-05 & 1.09e-03 & 1.54e-03 & \statsimilar 5.26e-04 & \statsimilar 5.86e-04 \\
    PLAyBOOK & 2.00e+00 & 4.76e-10 & 1.25e-04 & 1.81e-04 & 1.79e-05 & 2.51e-05 & 3.70e-03 & 4.35e-03 & \statsimilar 6.48e-04 & \statsimilar 8.50e-04 \\
    KB & 2.00e+00 & 1.19e-09 & 2.31e-03 & 3.32e-03 & 3.03e-05 & 3.28e-05 & 1.09e-03 & 1.41e-03 & \statsimilar 4.75e-02 & \statsimilar 5.92e-02 \\
    qEI & \best 1.12e-07 & \best 1.55e-07 & 5.83e-06 & 7.37e-06 & 7.84e-06 & 6.94e-06 & 9.49e-05 & 1.35e-04 & \statsimilar 1.82e-04 & \statsimilar 1.89e-04 \\
    TS & 2.00e+00 & 3.02e-08 & 4.58e-04 & 4.77e-04 & 1.94e-04 & 2.15e-04 & 1.28e-03 & 1.14e-03 & \statsimilar 1.78e-03 & \statsimilar 1.59e-03 \\
    \eSRS (0.1) & 2.00e+00 & 3.87e-11 & \best 6.07e-07 & \best 7.70e-07 & \best 1.51e-06 & \best 1.60e-06 & \statsimilar 1.07e-06 & \statsimilar 1.28e-06 & \statsimilar 6.65e-07 & \statsimilar 8.96e-07 \\
    \eSPF (0.1) & 2.00e+00 & 1.66e-12 & \statsimilar 1.20e-06 & \statsimilar 1.77e-06 & \statsimilar 1.91e-06 & \statsimilar 1.89e-06 & \statsimilar 4.21e-07 & \statsimilar 5.64e-07 & \statsimilar 3.27e-07 & \statsimilar 4.51e-07 \\
    \eExploit & 2.00e+00 & 1.18e-11 & \statsimilar 9.89e-07 & \statsimilar 1.28e-06 & \statsimilar 1.70e-06 & \statsimilar 1.83e-06 & \best 4.12e-07 & \best 4.66e-07 & \best 3.23e-07 & \best 4.62e-07 \\
\bottomrule
    \toprule
    \bfseries Method
    & \multicolumn{2}{c|}{\bfseries logSixHumpCamel (2)} 
    & \multicolumn{2}{c|}{\bfseries modHartman6 (6)} 
    & \multicolumn{2}{c|}{\bfseries logGSobol (10)} 
    & \multicolumn{2}{c|}{\bfseries logRosenbrock (10)} 
    & \multicolumn{2}{c}{\bfseries logStyblinskiTang (10)} \\ 
    & \multicolumn{1}{c}{Median} & \multicolumn{1}{c|}{MAD}
    & \multicolumn{1}{c}{Median} & \multicolumn{1}{c|}{MAD}
    & \multicolumn{1}{c}{Median} & \multicolumn{1}{c|}{MAD}
    & \multicolumn{1}{c}{Median} & \multicolumn{1}{c|}{MAD}
    & \multicolumn{1}{c}{Median} & \multicolumn{1}{c}{MAD}  \\ \midrule
    LP & 2.22e-01 & 2.59e-01 & \statsimilar 8.25e-04 & \statsimilar 1.04e-03 & \statsimilar 7.58e+00 & \statsimilar 1.96e+00 & 5.97e+00 & 8.36e-01 & 2.07e+00 & 3.76e-01 \\
    PLAyBOOK & 1.88e-01 & 2.41e-01 & \statsimilar 2.11e-03 & \statsimilar 2.56e-03 & 9.82e+00 & 1.45e+00 & 5.98e+00 & 1.48e+00 & 2.28e+00 & 2.89e-01 \\
    KB & 4.72e+00 & 1.35e+00 & \statsimilar 7.33e-03 & \statsimilar 7.64e-03 & \best 7.21e+00 & \best 1.65e+00 & \statsimilar 5.29e+00 & \statsimilar 2.19e+00 & 1.96e+00 & 3.08e-01 \\
    qEI & 1.49e-01 & 1.09e-01 & \statsimilar 1.44e-02 & \statsimilar 9.65e-03 & 9.84e+00 & 2.11e+00 & 7.94e+00 & 5.00e-01 & 2.28e+00 & 2.04e-01 \\
    TS & 1.15e+00 & 6.79e-01 & 3.94e-02 & 1.60e-02 & 1.03e+01 & 7.49e-01 & 8.48e+00 & 4.55e-01 & 2.87e+00 & 1.07e-01 \\
    \eSRS (0.1) & \statsimilar 1.38e-03 & \statsimilar 2.04e-03 & \best 3.08e-04 & \best 3.87e-04 & \statsimilar 8.07e+00 & \statsimilar 2.53e+00 & \statsimilar 5.03e+00 & \statsimilar 1.58e+00 & 2.05e+00 & 3.64e-01 \\
    \eSPF (0.1) & \best 3.90e-04 & \best 5.71e-04 & \statsimilar 3.09e-04 & \statsimilar 3.09e-04 & \statsimilar 8.19e+00 & \statsimilar 1.88e+00 & \statsimilar 4.61e+00 & \statsimilar 1.43e+00 & \best 1.81e+00 & \best 4.56e-01 \\
    \eExploit & \statsimilar 1.15e-03 & \statsimilar 1.70e-03 & \statsimilar 4.24e-04 & \statsimilar 4.90e-04 & \statsimilar 7.40e+00 & \statsimilar 2.23e+00 & \best 4.45e+00 & \best 1.44e+00 & \statsimilar 1.81e+00 & \statsimilar 3.55e-01 \\
\bottomrule
  \end{tabular}%
  }
\caption{Optimisation results with a batch size of $q=10$. Median absolute
distance from the optimum (left) and median absolute deviation from the median
(MAD, right) after 20 batches (200 function evaluations) across the 51 runs. 
The method with the lowest median performance is shown in dark grey, with those
with statistically equivalent performance shown in light grey.}
\label{tbl:synthetic_results}
\end{table*}
Figure~\ref{fig:conv_plots} shows the convergence plots of the various
algorithms on six test problems for $q \in \{5,10,20\}$. As 
might be expected, qEI tends to perform worse as $q$ increases,
which we suspect is linked to the dimensionality of the qEI acquisition
function being $d \times q$. For the 10-dimensional functions and $q=20$,
this requires global optimisation in a $200$-dimensional space, a far from trivial task.
The Thompson sampling method (TS) relies
upon there being sufficient  stochasticity in the surrogate model to
select batch locations that are well distributed in space. If there is too much
or too little variation in the locations selected, as appears to be the case in
these results, the batch will be selected in either locations with poor mean
values (too much variation) or all at the same location (too little variation).
Similar performance results for TS are also shown in \citep{alvi:asynclp}.

As shown in the convergence plots and Table~\ref{tbl:synthetic_results}, the
\eshotgun batch algorithms, \eSRS and \eSPF, both performed well across the
range of synthetic problems for all batch sizes. \eExploit, which always 
samples at and around the surrogate's best mean prediction, also performed well
across the majority of synthetic functions and was statistically equivalent to
\eSPF on all functions with a batch size of $q=10$. This indicates that fully
exploiting the model at each iteration and learning about the best mean
prediction's local landscape (via sampling its local neighbourhood) is a sound
strategy and mirrors the findings, that being greedy is good, of
\citet{rehbach:ei_pv} and \citet{death:egreedy} in the sequential setting.

Interestingly, on the modHartman6 function in particular 
(Figure~\ref{fig:conv_plots}, lower-left), $q=20$ led to better
median  $\fstar$ than for $q=5$, even though there were 4
times fewer batches (10 instead of 40) and therefore the surrogate model was
fitted far fewer times. This indicates that the model poorly estimated the
underlying function, thus misleading the optimisation process. However, the
expected trend prevails: an increase in $q$ generally led to a decrease in
the median $\fstar$ as well as a decrease in the rate of convergence. 

The acquisition penalisation-based methods, LP and PLAyBOOK, performed 
similarly, with LP slightly ahead of PLAyBOOK. The dominating factor
setting the   penalisation radii \eqref{eqn:ball_radius} in both methods is the Lipschitz
constant, which was 
estimated as being the largest value of $\lVert \nabla \mu(\bx) \rVert$ over 
the whole problem domain for LP and locally for PLAyBOOK. Since the global
Lipschitz constant will always be at least as large as a local one, it is 
perhaps unsurprising that LP performs better,  because a 
larger constant corresponds to a smaller radius of penalisation, meaning that
the batch points will be, on average, closer together and, therefore, more
similar to the better-performing \eshotgun batch methods.

Convergence plots for \eSRS and \eSPF have a well-defined step-like
appearance for several test functions, which is particularly visible in the plots
with larger batch sizes. This is a consequence of the batch selection process
because the first location in the batch $\xnext_1$  minimises the surrogate
model's mean function (recall Algorithm~\ref{alg:eshotgun}, 
line \ref{alg:eS:exploit_model}). It does, however, imply that the sequential
\egreedy strategy is driving the optimisation process as the subsequent
evaluations in the batch generally show little improvement over $f(\xnext_1)$. This also
means that the locations sampled around $\xnext_1$ are useful because they
improve the surrogate model accuracy, allowing the
surrogate's mean prediction to drive the optimisation.

\begin{figure}[t]
\includegraphics[width=\columnwidth, clip, trim={7 7 7 7}]{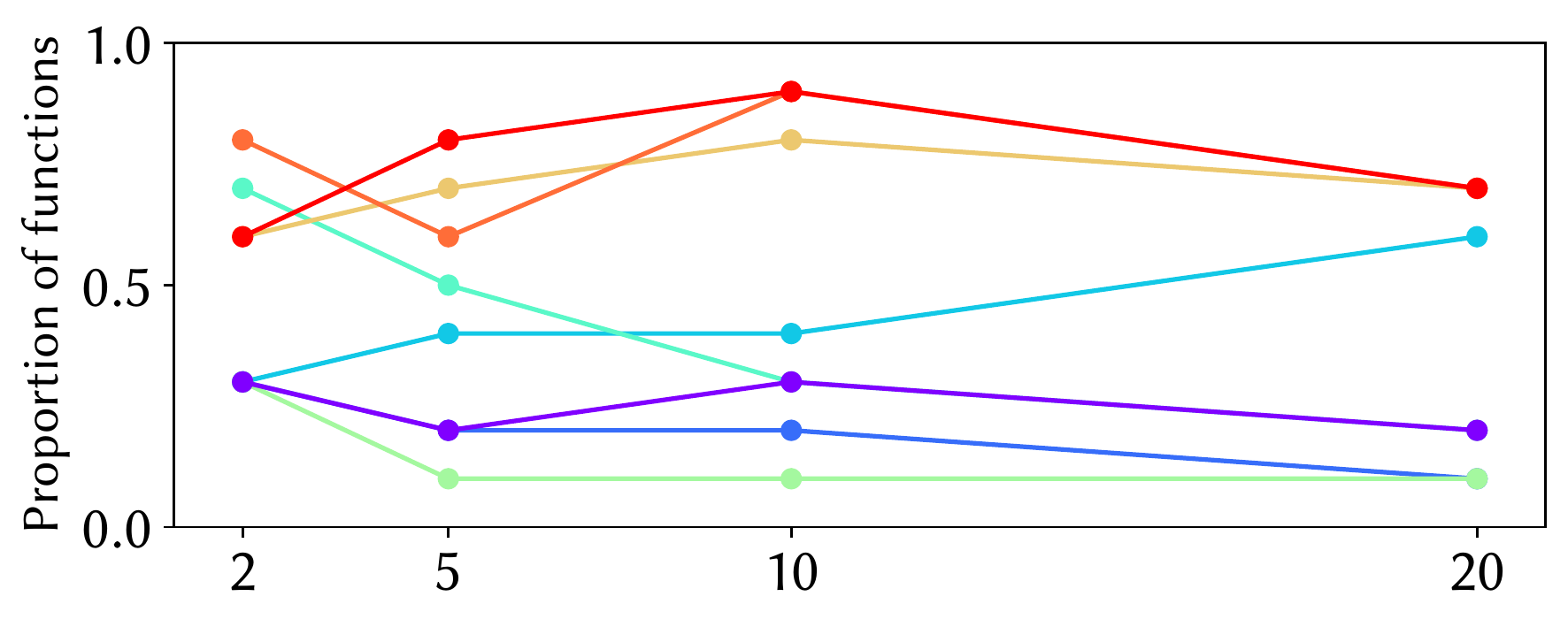}\\
\includegraphics[width=\columnwidth, clip, trim={10 10 10 10}]{figs/convergence_LEGEND}%
\caption{Synthetic function optimisation summary. Symbols correspond to the 
proportion of times that a method is best or statistically equivalent to the 
best method across the 10 synthetic functions for $q \in \{2,5,10,20\}$.}
\label{fig:summary_results}
\end{figure}
Figure~\ref{fig:summary_results} summarises the performance of each of the $7$
evaluated methods for the $4$ batch sizes. We note that the \eshotgun methods
are consistently the best or statistically indistinguishable from
the best performing methods across
the set of benchmark functions across all batch sizes. Interestingly, the older
Kriging Believer~\citep{ginsbourger:kb}, that penalises the 
surrogate model's variance around selected batch points, performed better 
than the newer, acquisition-based penalisers, particularly for larger batch
sizes. The increase in relative performance may be related to the particular
acquisition function used because, as shown in \citep{death:egreedy}, EI
weights improvements over the current $\fstar$ much more highly
than increases in variance. This may lead to the variance penalisation in KB
having a smaller radius of effect than the penalisation in EI-space by the 
LP and PLAyBOOK methods, resulting in KB sampling locations closer
together, in a more similar fashion to the \eshotgun methods.

As shown in Figure~\ref{fig:summary_results}, for the \eshotgun-based algorithms, there is little to differentiate overall 
between selecting a location at random from either the Pareto front (\eSPF) or
uniformly across the feasible space (\eSRS).
However, \eSPF appears to be marginally better on lower-dimensional functions,
most likely due to the surrogate model better describing the overall structure
of the modelled function. Conversely, \eSRS is slightly better on 
higher-dimensional functions because the modelled function with naturally be of
a poorer quality and therefore relying solely on it, without sufficient 
stochasticity, could hinder the optimisation process.

\begin{figure}[t]
\includegraphics[width=\columnwidth, clip, trim={7 7 7 7}]{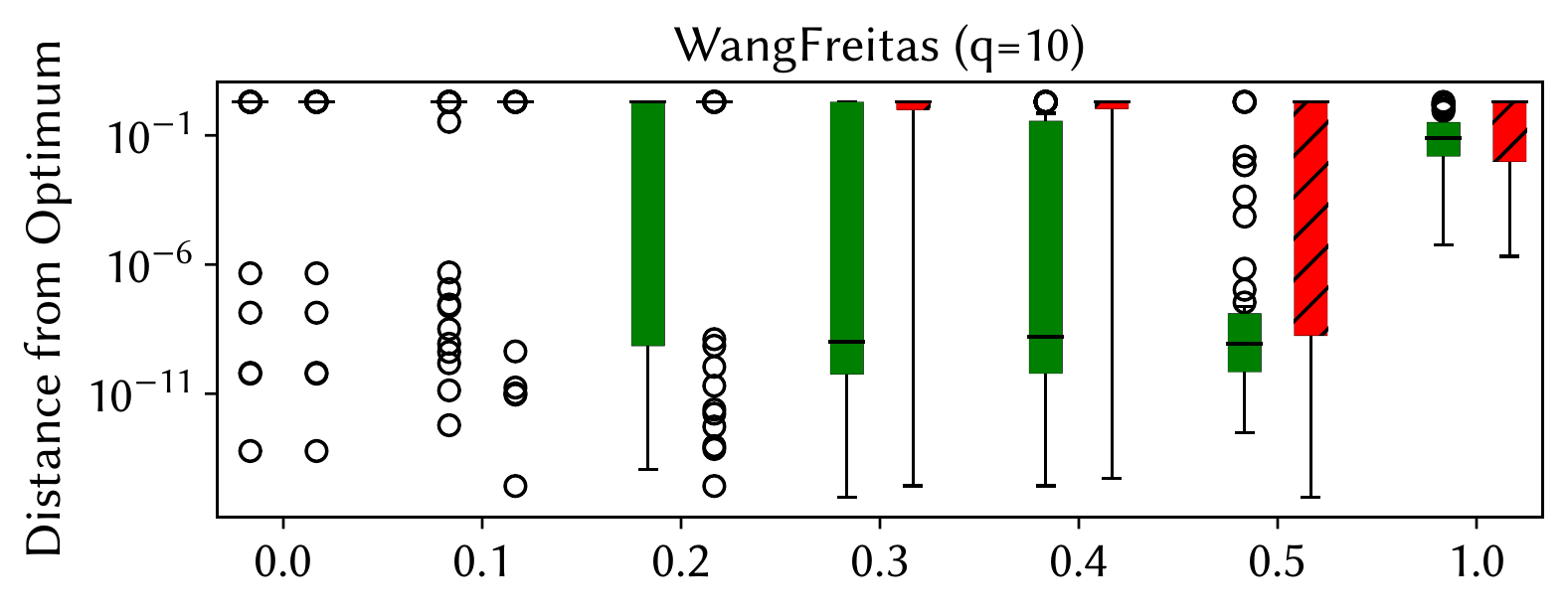}%
\caption{Distribution of $|\fmin - \fstar|$ after 200 function 
evaluations, taken over 51 runs, for \eSRS (green) and \eSPF (red, hatched) 
for different values of $\epsilon$ (horizontal axis) on the WangFreitas test problem with a batch
size of $q=10$.}
\label{fig:epsilon_compare_wangfreitas}
\end{figure}

Pure exploitation, \ie  $\epsilon=0$, the \eExploit method, leads to
state-of-the-art performance across for many problems and dimensionalities.
However, for problems in which
a large amount of exploration is needed in order to locate a deceptive
optimum, the \eshotgun methods with little exploration are unable to escape
local minima or expend enough of their optimisation budget exploring the
landscape. This  is particularly apparent on the WangFreitas problem
\citep{wangfreitas:theoreticalBO}, which has a large, shallow local minimum
and a narrow, deep global minimum surrounded by plateaus; see
\citep{wangfreitas:theoreticalBO} for a plot.
Figure~\ref{fig:epsilon_compare_wangfreitas} compares the
performance of the \eshotgun methods for different  $\epsilon$ on
this problem. Note how \eSRS (green) is able to more consistently find
the global optimum for smaller values of $\epsilon$, because, unlike \eSPF
(red), it is not constrained to only select non-dominated areas of decision
space. \eSPF, on the other hand, is consistently misled by the surrogate
model's incorrect estimation of the function and therefore fails to
correctly optimise the function even when $\epsilon = 0.5$.\looseness=-1

\subsection{Active Learning for Robot Pushing}
\label{sec:exps:robot}
Following \citep{wang:MES, jiang:nonmyopicbo,
death:egreedy}, we optimise the control parameters for two active learning
robot pushing problems \citep{wang:robots}; see \cite{death:egreedy} for
diagrams. In the first problem,
\pushfour, a robot is required to push an object towards an
unknown target location. It receives the object-target distance
once it has finished pushing. Its movement is constrained such that it
can only travel in the direction towards the object's initial location. The
parameters to be optimised are the robot's initial location, the orientation of
its hand and for how long it travels. Thus optimising the values of the
four parameters to reduce the final object-target distance can be cast as a minimisation problem. The object's
initial location is always set to be the centre of the domain
\citep{wang:MES, death:egreedy}, but the target location is changed in each
optimisation run, with these common across methods to ensure fair comparison.
The performance of an optimisation algorithm is thus averaged over problem
instances rather than the same function with different initialisations, as
with the synthetic functions previously.

Similarly, in the second problem \pusheight, two robots
push their own objects towards unknown targets, with the complication that
they  may  block each other's path. The 8 parameters controlling
both robots can be optimised to minimise the summed final object-target distances.
Initial object locations were fixed and target's positions were generated
randomly, while ensuring 
that each pair of target positions allowed each object to touch its target
without the objects overlapping. However, in some
problem instances it is not possible for both robots to push their objects
to their targets because the objects may be positioned such that the robots
need to cross each other's paths. In order to report the difference between
$\fstar$ and the true optimum we estimate the optimum of each problem instance
by randomly sampling in the feasible space with $10^5$ sets of robot parameters
and locally optimising the best 100 of these using L-BFGS-B. We therefore
report the difference between the algorithm's optimum and this estimated
global optimum.

\begin{figure}[t]
\includegraphics[width=\columnwidth, clip, trim={0 10 0 0}]{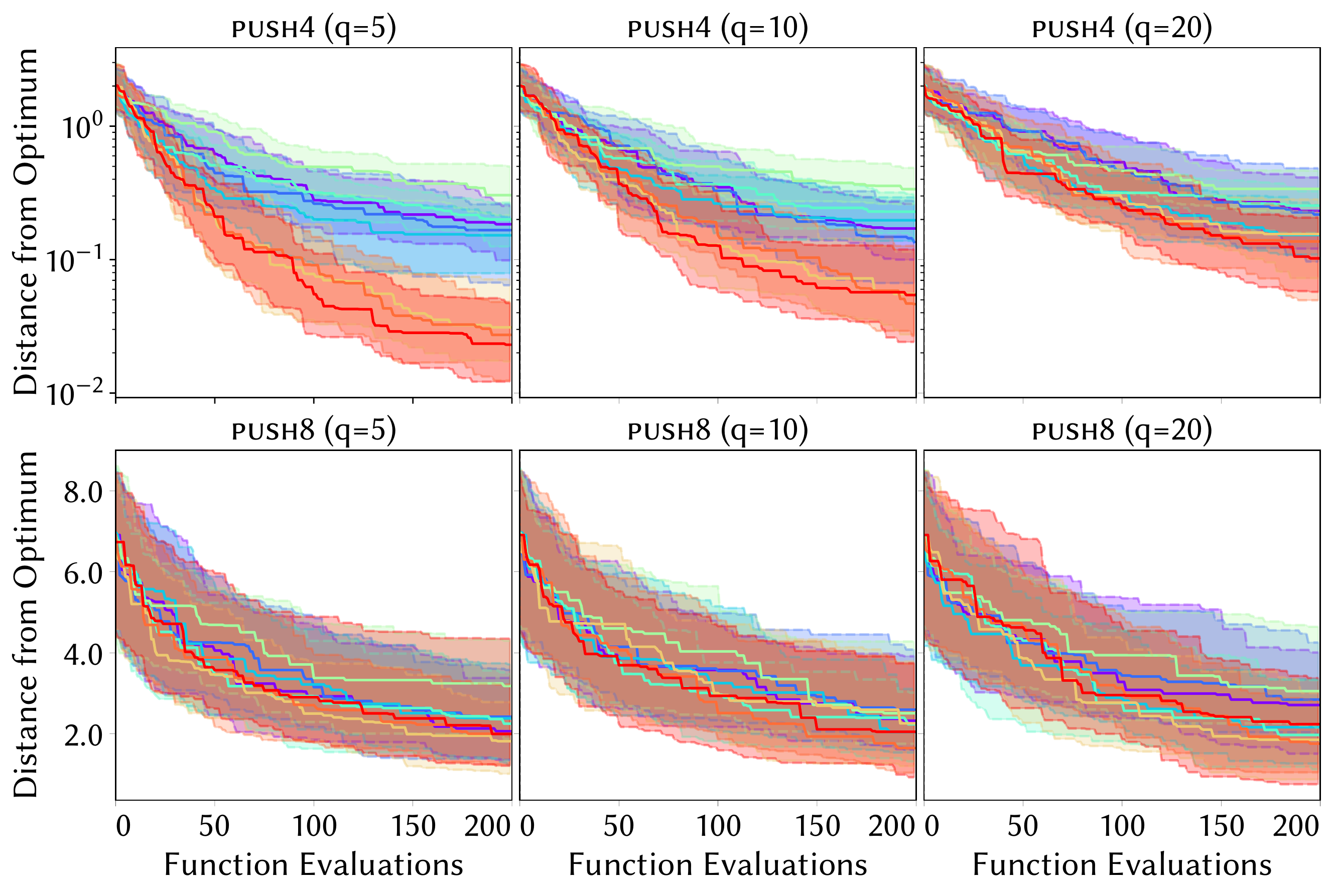}\\
\includegraphics[width=\columnwidth, clip, trim={10 10 10 10}]{figs/convergence_LEGEND}%
\caption{Convergence plots for the robot pushing problems (rows) over three
batch sizes $q \in \{5,10,20\}$ (columns). Each plot shows the median 
value of $\lvert \fstar - \fmin\rvert$,
with shading representing the interquartile range across the 51 runs.}
\label{fig:conv_plots_push}
\end{figure}
Figure~\ref{fig:conv_plots_push} shows the convergence %
for  $q \in \{5,10,20\}$. In  %
\pushfour the \eshotgun methods have statistically equivalent
performance, but outperform other methods for $q=5$ and $q=10$. KB also has 
statistically similar performance to the exploitative methods when $q=20$, echoing its
efficacy on the synthetic problems in which it also comparatively improved
with increasing $q$.

In the harder  \pusheight, all methods were
statistically equivalent with $q=10$, while TS and KB
were worse with $q=5$;  they were also worse, along with LP
for $q=20$. \eSRS and \eSPF consistently had the lowest
median fitnesses, although other techniques were statistically equivalent. Interestingly, and echoing the results on the synthetic modHartman6
function, \eSPF had a lower median fitness value for $q=10$ and $q=20$ than
for the smaller batch sizes. 

\subsection{Pipe Shape Optimisation}
\label{sec:exps:pipe}

We also evaluated the BBO methods on a real-world
computational fluid dynamics  (CFD) design problem. The PitzDaily test 
problem \citep{daniels:benchmark}, %
involves reducing the pressure loss along a  pipe of different inflow and outflow 
diameters by optimising the pipe's internal shape.  The optimisation aims to find the shape of the lower wall of the pipe
that minimises the pressure loss. The loss is evaluated by running a CFD
mesh generation and 
partial differential equation simulation of the two-dimensional flow. Each
function evaluation takes between 60\si{\second}
and 90\si{\second}, depending on mesh complexity.

The pipe's lower wall geometry is represented by a
Catmull-Clark sub-division curve, whose control points comprise the decision
variables. We use 5 control points,
resulting in a 10-dimensional decision vector. The control points are 
constrained to lie within  a polygon
and the initial locations used in the optimisation runs are uniformly
sampled in this constrained domain. 
Similarly, for the batch optimisation methods themselves, we take the naive
approach of rejection sampling when  optimising
acquisition functions for the KB, qEI and the \eshotgun approaches; we  do not consider locations that
violate the constraints for LP, PLAyBOOK and TS.

\begin{figure}[t]
\includegraphics[width=1\columnwidth, clip, trim={0 10 0 8}]{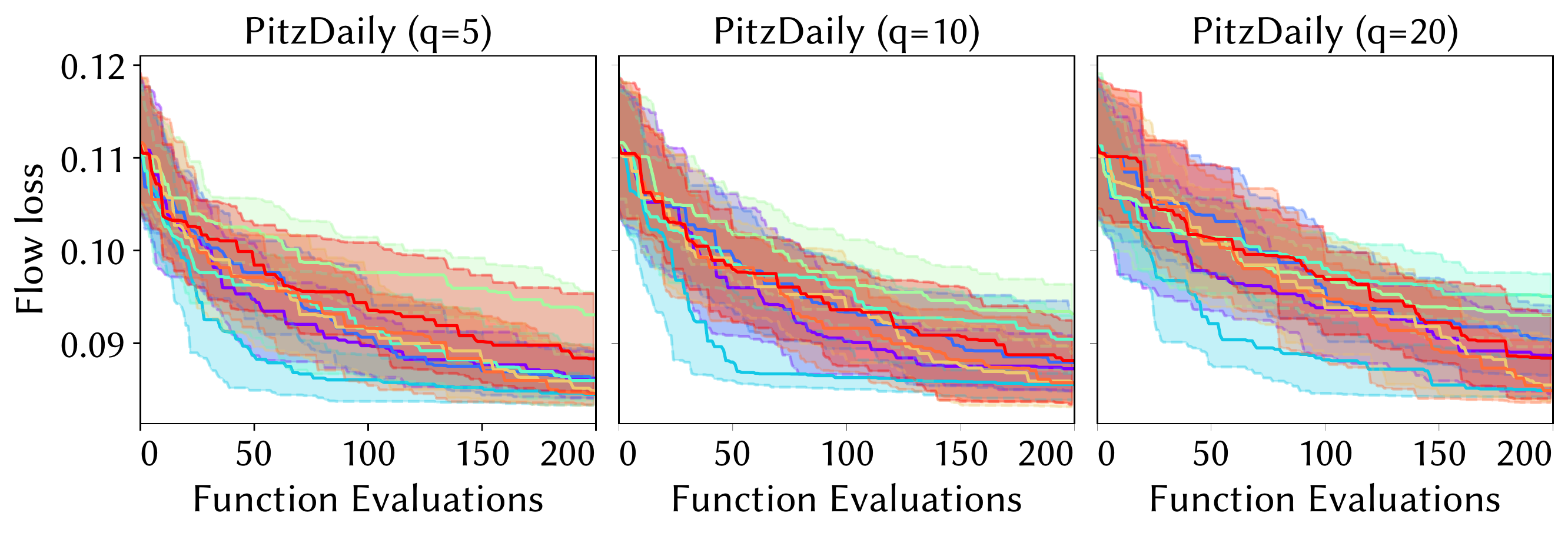}\\
\includegraphics[width=\columnwidth, clip, trim={10 10 10 10}]{figs/convergence_LEGEND}%
\caption{%
Convergence plots for the PitzDaily problem with $q \in \{5,10,20\}$. Each
plot shows the median best seen flow loss, with shading representing the IQR
across the 51 runs.}
\label{fig:conv_plots_pitzdaily}
\end{figure}
Convergence plots of the flow loss with $q \in \{5,10,20\}$ are shown in Figure~\ref{fig:conv_plots_pitzdaily}.
The Kriging Believer (KB)  consistently optimised the problem well,
although \eSRS, \eSPF, \eExploit and LP were statistically equivalent with
batch sizes $q=10$ and $q=20$. The rates at which KB reaches its best flow
losses, however, were superior to the other methods on the PitzDaily problem. In
the $q=10$ case, KB reaches close to its best flow losses after only 5 batches.
We note that all methods were able to discover pipe shape configurations that
led to flow losses that were better than $0.0903$ found by an adjoint (local, gradient-based)
optimisation method  \citep{nilsson:pitzdaily}, although TS, qEI,
and PLAYbOOK were not able to reach a median flow loss of lower than the 
adjoint solution for all batch sizes.

\section{Conclusions and Future Work}
\label{sec:conc}

Our novel \eshotgun method, which uses an \egreedy acquisition function to
select the first batch location and samples the remaining locations around
it, is both conceptually simple and computationally efficient because,
unlike many other batch methods, only one global optimisation run is needed
to select the batch locations. The method is competitive with
state-of-the-art BBO algorithms and better than them in several cases. We
attribute this to the exploitative nature of the first batch point selected
together with benefit of learning an accurate function model with the
remainder of the batch.

Pure exploitation ($\epsilon=0$: \eExploit) led to good performance on the majority of
problems because the surrogate model poorly estimates the true function,
particularly on higher-dimensional functions, thus inducing 
enough exploration. However, in the case of degenerate functions,
\eg WangFreitas, the surrogate model is too poor to optimise well, requiring a
larger $\epsilon$ to promote exploration. We have found that
$\epsilon=0.1$ works well.

Future research questions revolve around how best to select the locations of
the $q-1$ batch points. Although not described in detail here, we also investigated an alternative
\eshotgun approach of always selecting the first batch location as the 
surrogate model's best mean prediction and dividing the remaining samples into
two groups. One group, selected with probability $1-\epsilon$ for each 
location, was used identically to the greedy shotgun approach of sampling 
around the first batch location and the second group (probability $\epsilon$)
were randomly sampled in space or on the approximated Pareto front. This approach
gave similar results to \eshotgun.

One possible extension is use Latin hypercube sampling instead of drawing
random samples to better spread out the batch locations, making the
best use of each function evaluation. In addition, it may be beneficial to
tailor the sampling covariance to sample less/more densely in directions that
are flatter/steeper in decision space.

Lastly,  the function evaluation may take different times
depending on location and computational hardware, resulting in some
evaluations finishing before others. 
Current research, therefore, focuses on extending the \eshotgun method to 
asynchronous BBO.

\begin{acks} 
This work was supported by \grantsponsor{}{Innovate UK}{} [grant number \grantnum{}{104400}].
\end{acks}

\balance %
\bibliographystyle{ACM-Reference-Format}
\bibliography{ref}

\end{document}


\title{\eshotgun: \egreedy Batch Bayesian Optimisation}
\subtitle{Supplementary Material}

\author{George {De Ath}}
\email{g.de.ath@exeter.ac.uk}
\orcid{0000-0003-4909-0257}
\affiliation{%
  \department{Department of Computer Science}
  \institution{University of Exeter}
  \city{Exeter}
  \country{United Kingdom}
}

\author{Richard M. Everson}
\email{r.m.everson@exeter.ac.uk}
\orcid{0000-0002-3964-1150}
\affiliation{%
  \department{Department of Computer Science}
  \institution{University of Exeter}
  \city{Exeter}
  \country{United Kingdom}
}

\author{Jonathan E. Fieldsend}
\email{j.e.fieldsend@exeter.ac.uk}
\orcid{0000-0002-0683-2583}
\affiliation{%
  \department{Department of Computer Science}
  \institution{University of Exeter}
  \city{Exeter}
  \country{United Kingdom}
}

\author{Alma A. M. Rahat}
\email{a.a.m.rahat@swansea.ac.uk}
\orcid{0000-0002-5023-1371}
\affiliation{%
  \department{Department of Computer Science}
  \institution{Swansea University}
  \city{Swansea}
  \country{United Kingdom}
}

\renewcommand{\shortauthors}{De Ath et al.}

\maketitle

\section{Introduction}
In this supplementary paper, we show the formulae of each of the synthetic
functions used in this work and present the full batch optimisation results
and convergence plots with all batch sizes $q \in \{2,5,10,20\}$ for the
synthetic, robot pushing and pipe shape optimisation problems.

\section{Synthetic function details}
In the following section we give the formulae of each of the 10 synthetic
functions optimised in this work. Where functions have been modified from their
original form, we label the original functions as $g(\bx)$ and minimised
function as $f(\bx)$.

\subsection{WangFreitas}
\begin{align}
g(x) &=  2 \exp \left( -\frac{1}{2} \left(\frac{x - a}{\theta_1}\right)^2 \right)
       + 4 \exp \left( -\frac{1}{2} \left(\frac{x - b}{\theta_2}\right)^2 \right)
\label{eqn:wangfreitas_original} \\
f(x) &= -g(x),
\label{eqn:wangfreitas}
\end{align}
where $a=0.1$, $b=0.9$, $\theta_1=0.1$ and $\theta_2=0.01$.

\subsection{Branin}
\begin{equation}
f(\bx) = a\left(x_2 - b x_1^2 + c x_1 - r\right)^2 + s\left(1 - t\right) \cos\left(x_1\right) + s,
\label{eqn:branin}
\end{equation}
where $a=1$, $b=\tfrac{5.1}{4 \pi^2}$, $c=\tfrac{5}{\pi}$, $r=6$, $s=10$, 
$t=\tfrac{1}{8 \pi}$ and $x_i$ refers to the $i$-th element of vector $\bx$.

\subsection{BraninForrester}
\begin{equation}
f(\bx) = a(x_2 - b x_1^2 + c x_1 - r)^2 + s(1 - t) \cos(x_1) + s + 5 x_1,
\label{eqn:braninforrester}
\end{equation}
where $a=1$, $b=\tfrac{5.1}{4 \pi^2}$, $c=\tfrac{5}{\pi}$, $r=6$, $s=10$, 
and $t=\tfrac{1}{8 \pi}$.

\subsection{Cosines}
\begin{align}
g(\bx) =& 1 - \sum_{i=1}^2 \left[ \left(1.6 x_i - 0.5\right)^2
                              - 0.3 \cos \left(3 \pi \left(1.6 x_i - 0.5\right)\right) \right] \\
f(\bx) =& - g(\bx).
\label{eqn:cosines}
\end{align}

\subsection{logGoldsteinPrice}
\begin{align}
\begin{split}
g(\bx) = & \left(1 +             \left(x_1 + x_2 +1 \right)^2 \right.\\
         & \phantom{(1 + } \times \left. (19 - 14 x_1 + 3 x_1^2 - 14 x_2 + 6 x_1 x_2 + 3 x_2^2)\right) \\
         & \times \left(30 +             \left(2 x_1 - 3 x_2\right)^2 \right. \\
         & \phantom{\times (30 +} \times \left. (18 - 32 x_1 + 12 x_1^2 + 48 x_2 - 36 x_1 x_2 + 27 x_2^2)\right)
\end{split} \\
f(\bx) = & \log \left( g(\bx)\right).
\label{eqn:loggoldsteinprice}
\end{align}

\subsection{logSixHumpCamel}
\begin{align}
g(\bx) = & \left(4 - 2.1 x_1^2 + \frac{x_1^4}{3}\right) x_1^2 
            + x_1 x_2 + \left(-4 + 4 x_2^2\right) x_2^2 \\
f(\bx) = & \log \left( g(\bx)  + a + b \right),
\label{eqn:logsixhumpcamel}
\end{align}
where $a = 1.0316$ and $b = 10^{-4}$. Note that because $g(\bx)$ has a minimum
value of $-1.0316$, we add $a$ plus a small constant ($b$) to avoid taking the
logarithm of a negative number; this does not change the function's landscape.

\subsection{modHartman6}
\begin{align}
g(\bx) = & -\sum_{i=1}^4 \alpha_i \exp \left( -\sum_{j=1}^6 A_{i j} \left(x_j - P_{i j} \right)^2 \right) \\
f(\bx) = & - \log \left( - g(\bx) \right)
\label{eqn:modhartman6}
\end{align}
where
\begin{align}
\mathbf{\alpha} = & \left( 1.0, 1.2, 3.0, 3.2 \right)^T \\
\mathbf{A} = & \left(\begin{array}{cccccc} 10   & 3   & 17   & 3.50 & 1.7 & 8 \\ 
                                           0.05 & 10  & 17   & 0.1  & 8   & 14 \\ 
                                           3    & 3.5 & 1.7  & 10   & 17  & 8 \\ 
                                           17   & 8   & 0.05 & 10   & 0.1 & 14
                     \end{array}\right) \\
\mathbf{P} = & 10^{-4} \left(\begin{array}{cccccc} 1312 & 1696 & 5569 &  124 & 8283 & 5886 \\ 
                                                   2329 & 4135 & 8307 & 3736 & 1004 & 9991 \\ 
                                                   2348 & 1451 & 3522 & 2883 & 3047 & 6650 \\ 
                                                   4047 & 8828 & 8732 & 5743 & 1091 &  381 
                             \end{array}\right).
\label{eqn:modhartman6_variables}
\end{align}

\subsection{logGSobol}
\begin{align}
g(\bx) = & \prod_{i=1}^D \frac{4 x_i - 1}{2} \\
f(\bx) = & \log \left( g(\bx) \right),
\label{eqn:loggsobol}
\end{align}
where $a = 1$ and $D = 10$.

\subsection{logRosenbrock}
\begin{align}
g(\bx) = & \sum_{i=1}^{D-1} \left[ 100 \left(x_{i+1} - x_i^2\right)^2 
                                   + \left(x_i - 1\right)^2 \right] \\
f(\bx) = & \log \left( g(\bx) + 0.5 \right),
\label{eqn:logrosenbrock}
\end{align}
where $D = 10$. Note, similarly to logSixHumpCamel, because $g(\bx)$ has a 
minimum value of $0$, we add a value to ensure it is always positive.

\subsection{logStyblinskiTang}
\begin{align}
g(\bx) = & \frac{1}{2} \sum_{i=1}^D \left(x_i^4 - 16 x_i^2 + 5 x_i\right) \\
f(\bx) = & \log \left( g(\bx) + 40D \right),
\label{eqn:logstyblinskitang}
\end{align}
where $D = 10$. Since $g(\bx)$ has a minimum value of $-39.16599D$, 
we add $40D$ to it to ensure it is always positive.

\section{Additional Results}
In the following we display the full set of results for the experimental
evaluations carried out in this paper. In Section~\ref{sec:synthetic_functions}
we display tabulated results for the synthetic functions with batch sizes
$q \in \{2,5,20\}$ and show convergence plots for all batch sizes. Similarly,
in Sections~\ref{sec:robot_pushing}~and~\ref{sec:pitzdaily} we display
tabulated results and convergence plots for the robot pushing and PitzDaily
real-world test problems.

\subsection{Synthetic functions}
\label{sec:synthetic_functions}
  \begin{table}[ht]
  \setlength{\tabcolsep}{2pt}
  \sisetup{table-format=1.2e-1,table-number-alignment=center}
  \resizebox{1\textwidth}{!}{%
  \begin{tabular}{l | SS| SS| SS| SS| SS}
    \toprule
    \bfseries Method
    & \multicolumn{2}{c|}{\bfseries WangFreitas (1)} 
    & \multicolumn{2}{c|}{\bfseries BraninForrester (2)} 
    & \multicolumn{2}{c|}{\bfseries Branin (2)} 
    & \multicolumn{2}{c|}{\bfseries Cosines (2)} 
    & \multicolumn{2}{c}{\bfseries logGoldsteinPrice (2)} \\ 
    & \multicolumn{1}{c}{Median} & \multicolumn{1}{c|}{MAD}
    & \multicolumn{1}{c}{Median} & \multicolumn{1}{c|}{MAD}
    & \multicolumn{1}{c}{Median} & \multicolumn{1}{c|}{MAD}
    & \multicolumn{1}{c}{Median} & \multicolumn{1}{c|}{MAD}
    & \multicolumn{1}{c}{Median} & \multicolumn{1}{c}{MAD}  \\ \midrule
    LP & 2.00e+00 & 2.15e-08 & 4.23e-05 & 5.66e-05 & 3.24e-05 & 4.30e-05 & \statsimilar 1.73e-04 & \statsimilar 2.44e-04 & \statsimilar 5.56e-05 & \statsimilar 7.75e-05 \\
    PLAyBOOK & 2.00e+00 & 3.73e-09 & 3.16e-05 & 4.28e-05 & 3.07e-05 & 4.06e-05 & \statsimilar 1.33e-04 & \statsimilar 1.76e-04 & \statsimilar 3.71e-05 & \statsimilar 3.73e-05 \\
    KB & 2.00e+00 & 7.68e-09 & 1.07e-04 & 1.47e-04 & 2.42e-05 & 2.99e-05 & \statsimilar 6.54e-04 & \statsimilar 6.64e-04 & \statsimilar 4.02e-02 & \statsimilar 4.70e-02 \\
    qEI & \statsimilar 2.00e+00 & \statsimilar 9.01e-11 & \best 2.47e-07 & \best 3.49e-07 & \statsimilar 3.26e-06 & \statsimilar 3.94e-06 & \statsimilar 1.09e-05 & \statsimilar 1.39e-05 & \statsimilar 9.33e-05 & \statsimilar 1.31e-04 \\
    TS & 2.00e+00 & 5.68e-09 & 1.14e-04 & 1.17e-04 & 9.04e-05 & 9.53e-05 & \statsimilar 2.69e-04 & \statsimilar 3.23e-04 & \statsimilar 2.04e-04 & \statsimilar 2.23e-04 \\
    \eSRS (0.1) & \statsimilar 2.00e+00 & \statsimilar 5.66e-08 & 6.80e-06 & 1.00e-05 & \statsimilar 3.21e-06 & \statsimilar 3.90e-06 & \statsimilar 3.56e-06 & \statsimilar 5.11e-06 & 9.71e-07 & 1.42e-06 \\
    \eSPF (0.1) & \best 2.00e+00 & \best 3.70e-13 & 2.07e-06 & 2.95e-06 & \best 2.43e-06 & \best 2.45e-06 & \statsimilar 4.26e-06 & \statsimilar 6.12e-06 & \best 9.39e-08 & \best 1.31e-07 \\
    \eExploit & 2.00e+00 & 2.52e-09 & 7.58e-06 & 1.09e-05 & \statsimilar 2.46e-06 & \statsimilar 3.01e-06 & \best 1.85e-06 & \best 2.37e-06 & \statsimilar 1.15e-07 & \statsimilar 1.66e-07 \\
\bottomrule
    \toprule
    \bfseries Method
    & \multicolumn{2}{c|}{\bfseries logSixHumpCamel (2)} 
    & \multicolumn{2}{c|}{\bfseries modHartman6 (6)} 
    & \multicolumn{2}{c|}{\bfseries logGSobol (10)} 
    & \multicolumn{2}{c|}{\bfseries logRosenbrock (10)} 
    & \multicolumn{2}{c}{\bfseries logStyblinskiTang (10)} \\ 
    & \multicolumn{1}{c}{Median} & \multicolumn{1}{c|}{MAD}
    & \multicolumn{1}{c}{Median} & \multicolumn{1}{c|}{MAD}
    & \multicolumn{1}{c}{Median} & \multicolumn{1}{c|}{MAD}
    & \multicolumn{1}{c}{Median} & \multicolumn{1}{c|}{MAD}
    & \multicolumn{1}{c}{Median} & \multicolumn{1}{c}{MAD}  \\ \midrule
    LP & 1.29e-01 & 1.44e-01 & \statsimilar 5.90e-04 & \statsimilar 6.04e-04 & 7.07e+00 & 1.84e+00 & 5.96e+00 & 1.69e+00 & 2.17e+00 & 4.35e-01 \\
    PLAyBOOK & 1.12e-01 & 1.22e-01 & \statsimilar 5.22e-04 & \statsimilar 5.69e-04 & 8.01e+00 & 1.84e+00 & 5.44e+00 & 1.42e+00 & 2.11e+00 & 2.55e-01 \\
    KB & 4.86e+00 & 1.07e+00 & \statsimilar 4.53e-03 & \statsimilar 2.89e-03 & 7.64e+00 & 1.21e+00 & 4.55e+00 & 1.42e+00 & 2.08e+00 & 3.34e-01 \\
    qEI & 1.78e-02 & 1.92e-02 & \statsimilar 2.72e-03 & \statsimilar 2.27e-03 & \best 6.83e+00 & \best 1.57e+00 & 5.36e+00 & 1.57e+00 & 2.07e+00 & 2.70e-01 \\
    TS & 3.28e-01 & 3.05e-01 & \statsimilar 2.40e-02 & \statsimilar 1.04e-02 & 9.84e+00 & 9.55e-01 & 8.08e+00 & 3.93e-01 & 2.85e+00 & 1.52e-01 \\
    \eSRS (0.1) & \best 7.64e-05 & \best 8.01e-05 & \best 4.73e-04 & \best 6.02e-04 & 9.43e+00 & 3.01e+00 & \best 3.62e+00 & \best 1.01e+00 & 1.81e+00 & 3.65e-01 \\
    \eSPF (0.1) & \statsimilar 1.13e-04 & \statsimilar 1.26e-04 & \statsimilar 6.67e-04 & \statsimilar 8.69e-04 & 8.46e+00 & 1.92e+00 & \statsimilar 4.91e+00 & \statsimilar 2.61e+00 & \best 1.81e+00 & \best 4.67e-01 \\
    \eExploit & 1.61e-04 & 2.28e-04 & \statsimilar 8.30e-04 & \statsimilar 8.45e-04 & 9.24e+00 & 2.10e+00 & \statsimilar 3.68e+00 & \statsimilar 1.04e+00 & \statsimilar 1.81e+00 & \statsimilar 3.64e-01 \\
\bottomrule
  \end{tabular}%
  }%
\caption{Optimisation results with a batch size of $q=2$. Median absolute
distance from the optimum (left) and median absolute deviation from the median
(MAD, right) after 100 batches (200 function evaluations) across the 51 runs. 
The method with the lowest median performance is shown in dark grey, with those
with statistically equivalent performance shown in light grey.}
  \label{tbl:synthetic_results_bs2}
  \end{table}
  
  \begin{table}[ht]
  \setlength{\tabcolsep}{2pt}
  \sisetup{table-format=1.2e-1,table-number-alignment=center}
  \resizebox{1\textwidth}{!}{%
  \begin{tabular}{l | SS| SS| SS| SS| SS}
    \toprule
    \bfseries Method
    & \multicolumn{2}{c|}{\bfseries WangFreitas (1)} 
    & \multicolumn{2}{c|}{\bfseries BraninForrester (2)} 
    & \multicolumn{2}{c|}{\bfseries Branin (2)} 
    & \multicolumn{2}{c|}{\bfseries Cosines (2)} 
    & \multicolumn{2}{c}{\bfseries logGoldsteinPrice (2)} \\ 
    & \multicolumn{1}{c}{Median} & \multicolumn{1}{c|}{MAD}
    & \multicolumn{1}{c}{Median} & \multicolumn{1}{c|}{MAD}
    & \multicolumn{1}{c}{Median} & \multicolumn{1}{c|}{MAD}
    & \multicolumn{1}{c}{Median} & \multicolumn{1}{c|}{MAD}
    & \multicolumn{1}{c}{Median} & \multicolumn{1}{c}{MAD}  \\ \midrule
    LP & 2.00e+00 & 9.13e-10 & 5.39e-05 & 7.43e-05 & 1.17e-05 & 1.49e-05 & 5.16e-04 & 7.04e-04 & \statsimilar 1.01e-04 & \statsimilar 1.16e-04 \\
    PLAyBOOK & 2.00e+00 & 3.72e-10 & 2.48e-05 & 3.66e-05 & 2.15e-05 & 3.02e-05 & 2.97e-03 & 4.18e-03 & \statsimilar 4.93e-04 & \statsimilar 6.12e-04 \\
    KB & 2.00e+00 & 2.96e-09 & 5.19e-04 & 7.32e-04 & 3.03e-05 & 2.65e-05 & 1.09e-03 & 1.31e-03 & \statsimilar 4.55e-02 & \statsimilar 5.87e-02 \\
    qEI & \best 1.18e-08 & \best 1.69e-08 & \statsimilar 9.61e-07 & \statsimilar 9.95e-07 & \statsimilar 2.30e-06 & \statsimilar 1.90e-06 & 1.31e-04 & 1.41e-04 & \statsimilar 8.41e-05 & \statsimilar 1.00e-04 \\
    TS & 2.00e+00 & 1.32e-08 & 2.35e-04 & 2.52e-04 & 1.21e-04 & 1.11e-04 & 4.71e-04 & 5.93e-04 & \statsimilar 6.26e-04 & \statsimilar 5.77e-04 \\
    \eSRS (0.1) & 2.00e+00 & 6.66e-11 & \statsimilar 1.02e-06 & \statsimilar 1.47e-06 & \statsimilar 1.53e-06 & \statsimilar 1.66e-06 & 1.77e-06 & 2.28e-06 & \statsimilar 1.81e-06 & \statsimilar 2.61e-06 \\
    \eSPF (0.1) & 2.00e+00 & 3.59e-13 & \best 9.26e-07 & \best 1.35e-06 & \best 1.31e-06 & \best 1.23e-06 & 1.42e-06 & 1.52e-06 & \statsimilar 3.82e-07 & \statsimilar 5.65e-07 \\
    \eExploit & 2.00e+00 & 4.09e-11 & \statsimilar 2.59e-06 & \statsimilar 3.80e-06 & \statsimilar 2.52e-06 & \statsimilar 2.73e-06 & \best 8.21e-07 & \best 1.01e-06 & \best 3.60e-07 & \best 5.09e-07 \\
\bottomrule
    \toprule
    \bfseries Method
    & \multicolumn{2}{c|}{\bfseries logSixHumpCamel (2)} 
    & \multicolumn{2}{c|}{\bfseries modHartman6 (6)} 
    & \multicolumn{2}{c|}{\bfseries logGSobol (10)} 
    & \multicolumn{2}{c|}{\bfseries logRosenbrock (10)} 
    & \multicolumn{2}{c}{\bfseries logStyblinskiTang (10)} \\ 
    & \multicolumn{1}{c}{Median} & \multicolumn{1}{c|}{MAD}
    & \multicolumn{1}{c}{Median} & \multicolumn{1}{c|}{MAD}
    & \multicolumn{1}{c}{Median} & \multicolumn{1}{c|}{MAD}
    & \multicolumn{1}{c}{Median} & \multicolumn{1}{c|}{MAD}
    & \multicolumn{1}{c}{Median} & \multicolumn{1}{c}{MAD}  \\ \midrule
    LP & 1.58e-01 & 2.14e-01 & \statsimilar 8.77e-04 & \statsimilar 9.82e-04 & 8.16e+00 & 2.07e+00 & 5.84e+00 & 1.14e+00 & 2.18e+00 & 3.77e-01 \\
    PLAyBOOK & 1.55e-01 & 1.79e-01 & \statsimilar 7.03e-04 & \statsimilar 7.37e-04 & 8.53e+00 & 2.05e+00 & 5.69e+00 & 1.05e+00 & 2.26e+00 & 3.15e-01 \\
    KB & 4.69e+00 & 1.14e+00 & \statsimilar 6.10e-03 & \statsimilar 6.06e-03 & \best 7.10e+00 & \best 1.62e+00 & \statsimilar 4.68e+00 & \statsimilar 1.67e+00 & 2.09e+00 & 3.39e-01 \\
    qEI & 5.04e-02 & 5.54e-02 & \statsimilar 3.77e-03 & \statsimilar 2.73e-03 & 8.48e+00 & 2.22e+00 & 7.06e+00 & 1.10e+00 & 2.12e+00 & 3.58e-01 \\
    TS & 5.66e-01 & 4.71e-01 & 2.87e-02 & 1.04e-02 & 1.03e+01 & 8.36e-01 & 8.11e+00 & 4.52e-01 & 2.85e+00 & 1.08e-01 \\
    \eSRS (0.1) & \best 5.05e-05 & \best 5.35e-05 & \best 2.85e-04 & \best 2.85e-04 & 8.27e+00 & 2.08e+00 & \statsimilar 5.60e+00 & \statsimilar 1.94e+00 & \statsimilar 1.81e+00 & \statsimilar 3.63e-01 \\
    \eSPF (0.1) & 2.28e-04 & 3.17e-04 & \statsimilar 4.03e-04 & \statsimilar 5.00e-04 & 7.83e+00 & 2.24e+00 & \statsimilar 3.99e+00 & \statsimilar 1.45e+00 & \best 1.81e+00 & \best 4.84e-01 \\
    \eExploit & 3.95e-04 & 5.73e-04 & \statsimilar 4.51e-04 & \statsimilar 5.05e-04 & \statsimilar 7.86e+00 & \statsimilar 1.85e+00 & \best 3.97e+00 & \best 1.39e+00 & \statsimilar 1.81e+00 & \statsimilar 3.63e-01 \\
\bottomrule
  \end{tabular}%
  }%
\caption{Optimisation results with a batch size of $q=5$. Median absolute
distance from the optimum (left) and median absolute deviation from the median
(MAD, right) after 40 batches (200 function evaluations) across the 51 runs. 
The method with the lowest median performance is shown in dark grey, with those
with statistically equivalent performance shown in light grey.}
  \label{tbl:synthetic_results_bs5}
  \end{table}
  
  \begin{table}[ht]
  \setlength{\tabcolsep}{2pt}
  \sisetup{table-format=1.2e-1,table-number-alignment=center}
  \resizebox{1\textwidth}{!}{%
  \begin{tabular}{l | SS| SS| SS| SS| SS}
    \toprule
    \bfseries Method
    & \multicolumn{2}{c|}{\bfseries WangFreitas (1)} 
    & \multicolumn{2}{c|}{\bfseries BraninForrester (2)} 
    & \multicolumn{2}{c|}{\bfseries Branin (2)} 
    & \multicolumn{2}{c|}{\bfseries Cosines (2)} 
    & \multicolumn{2}{c}{\bfseries logGoldsteinPrice (2)} \\ 
    & \multicolumn{1}{c}{Median} & \multicolumn{1}{c|}{MAD}
    & \multicolumn{1}{c}{Median} & \multicolumn{1}{c|}{MAD}
    & \multicolumn{1}{c}{Median} & \multicolumn{1}{c|}{MAD}
    & \multicolumn{1}{c}{Median} & \multicolumn{1}{c|}{MAD}
    & \multicolumn{1}{c}{Median} & \multicolumn{1}{c}{MAD}  \\ \midrule
    LP & 2.00e+00 & 1.10e-09 & 2.80e-06 & 3.93e-06 & 3.81e-06 & 4.89e-06 & 2.25e-02 & 2.90e-02 & \statsimilar 1.90e-03 & \statsimilar 2.70e-03 \\
    PLAyBOOK & 8.28e-05 & 1.23e-04 & 9.99e-06 & 1.45e-05 & 3.36e-06 & 3.97e-06 & 1.07e-01 & 1.58e-01 & \statsimilar 1.68e-03 & \statsimilar 2.44e-03 \\
    KB & 8.58e-07 & 1.26e-06 & \best 2.96e-07 & \best 3.59e-07 & \best 1.14e-06 & \best 1.09e-06 & 2.83e-05 & 4.13e-05 & \statsimilar 2.27e-03 & \statsimilar 2.83e-03 \\
    qEI & \best 1.79e-07 & \best 2.56e-07 & 6.18e-05 & 5.57e-05 & 4.19e-05 & 4.14e-05 & 1.21e-03 & 1.65e-03 & \statsimilar 3.57e-04 & \statsimilar 3.86e-04 \\
    TS & 2.00e+00 & 1.60e-07 & 9.09e-04 & 9.60e-04 & 6.50e-04 & 5.91e-04 & 2.48e-03 & 1.96e-03 & \statsimilar 2.96e-03 & \statsimilar 3.03e-03 \\
    \eSRS (0.1) & 2.00e+00 & 2.07e-08 & 8.87e-07 & 1.24e-06 & 1.73e-06 & 1.90e-06 & \statsimilar 7.06e-07 & \statsimilar 1.01e-06 & \statsimilar 3.02e-06 & \statsimilar 4.43e-06 \\
    \eSPF (0.1) & 2.00e+00 & 2.15e-11 & 1.49e-06 & 2.16e-06 & 2.24e-06 & 2.49e-06 & \statsimilar 1.14e-06 & \statsimilar 1.55e-06 & \best 7.42e-07 & \best 9.61e-07 \\
    \eExploit & 2.00e+00 & 4.01e-11 & 7.46e-07 & 8.54e-07 & 1.69e-06 & 1.72e-06 & \best 6.06e-07 & \best 8.44e-07 & \statsimilar 9.03e-07 & \statsimilar 1.31e-06 \\
\bottomrule
    \toprule
    \bfseries Method
    & \multicolumn{2}{c|}{\bfseries logSixHumpCamel (2)} 
    & \multicolumn{2}{c|}{\bfseries modHartman6 (6)} 
    & \multicolumn{2}{c|}{\bfseries logGSobol (10)} 
    & \multicolumn{2}{c|}{\bfseries logRosenbrock (10)} 
    & \multicolumn{2}{c}{\bfseries logStyblinskiTang (10)} \\ 
    & \multicolumn{1}{c}{Median} & \multicolumn{1}{c|}{MAD}
    & \multicolumn{1}{c}{Median} & \multicolumn{1}{c|}{MAD}
    & \multicolumn{1}{c}{Median} & \multicolumn{1}{c|}{MAD}
    & \multicolumn{1}{c}{Median} & \multicolumn{1}{c|}{MAD}
    & \multicolumn{1}{c}{Median} & \multicolumn{1}{c}{MAD}  \\ \midrule
    LP & 5.35e-01 & 5.93e-01 & 3.24e-03 & 4.31e-03 & \statsimilar 8.32e+00 & \statsimilar 1.58e+00 & 6.68e+00 & 8.03e-01 & 2.13e+00 & 3.72e-01 \\
    PLAyBOOK & 6.35e-01 & 6.76e-01 & 2.40e-02 & 3.14e-02 & 1.00e+01 & 2.37e+00 & 6.29e+00 & 1.32e+00 & 2.25e+00 & 2.92e-01 \\
    KB & 1.65e+00 & 1.36e+00 & \statsimilar 2.51e-04 & \statsimilar 2.92e-04 & \statsimilar 8.17e+00 & \statsimilar 1.83e+00 & \statsimilar 5.49e+00 & \statsimilar 1.56e+00 & 2.11e+00 & 3.74e-01 \\
    qEI & 4.98e-01 & 3.87e-01 & 2.79e-02 & 1.73e-02 & 1.06e+01 & 1.77e+00 & 7.97e+00 & 6.08e-01 & 2.34e+00 & 2.06e-01 \\
    TS & 1.19e+00 & 9.82e-01 & 5.04e-02 & 1.47e-02 & 1.07e+01 & 1.21e+00 & 8.63e+00 & 4.04e-01 & 2.84e+00 & 1.47e-01 \\
    \eSRS (0.1) & \best 4.44e-03 & \best 6.56e-03 & \statsimilar 2.08e-04 & \statsimilar 2.62e-04 & \statsimilar 8.71e+00 & \statsimilar 2.45e+00 & \statsimilar 5.34e+00 & \statsimilar 1.26e+00 & \best 1.89e+00 & \best 2.93e-01 \\
    \eSPF (0.1) & \statsimilar 1.04e-02 & \statsimilar 1.53e-02 & \best 1.57e-04 & \best 1.79e-04 & \best 8.16e+00 & \best 3.04e+00 & \statsimilar 5.60e+00 & \statsimilar 1.20e+00 & \statsimilar 2.05e+00 & \statsimilar 3.48e-01 \\
    \eExploit & \statsimilar 2.55e-02 & \statsimilar 3.77e-02 & \statsimilar 1.87e-04 & \statsimilar 2.24e-04 & \statsimilar 8.90e+00 & \statsimilar 2.94e+00 & \best 5.00e+00 & \best 1.35e+00 & \statsimilar 2.05e+00 & \statsimilar 3.45e-01 \\
\bottomrule
  \end{tabular}%
  }%
\caption{Optimisation results with a batch size of $q=20$. Median absolute
distance from the optimum (left) and median absolute deviation from the median
(MAD, right) after 10 batches (200 function evaluations) across the 51 runs. 
The method with the lowest median performance is shown in dark grey, with those
with statistically equivalent performance shown in light grey.}
  \label{tbl:synthetic_results_bs20}
  \end{table}

\begin{figure}[ht!]
\includegraphics[width=\columnwidth, clip, trim={0 0 0 0}]{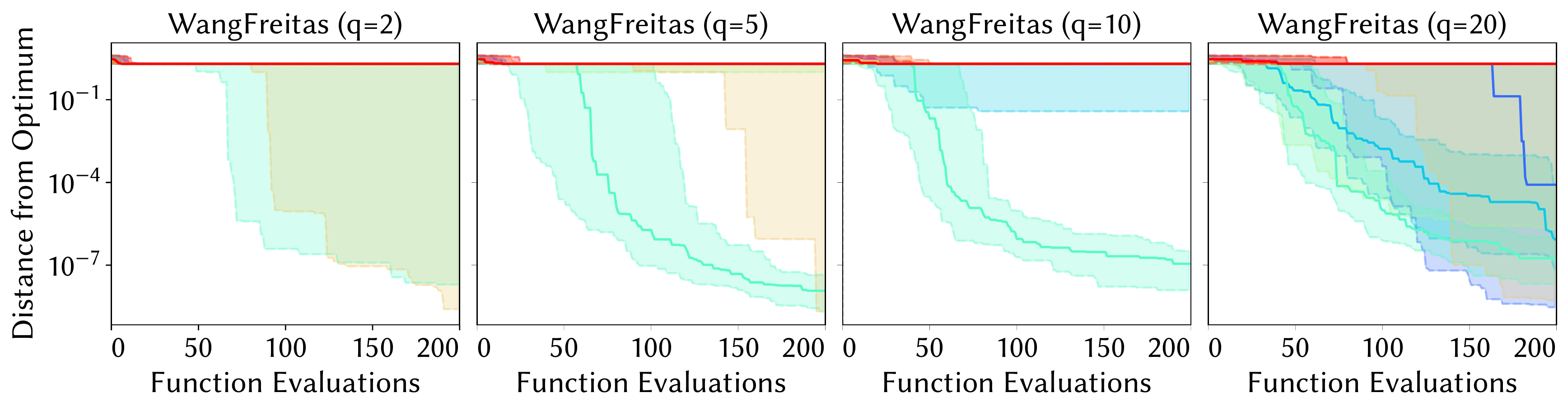}\\
\includegraphics[width=\columnwidth, clip, trim={0 0 0 0}]{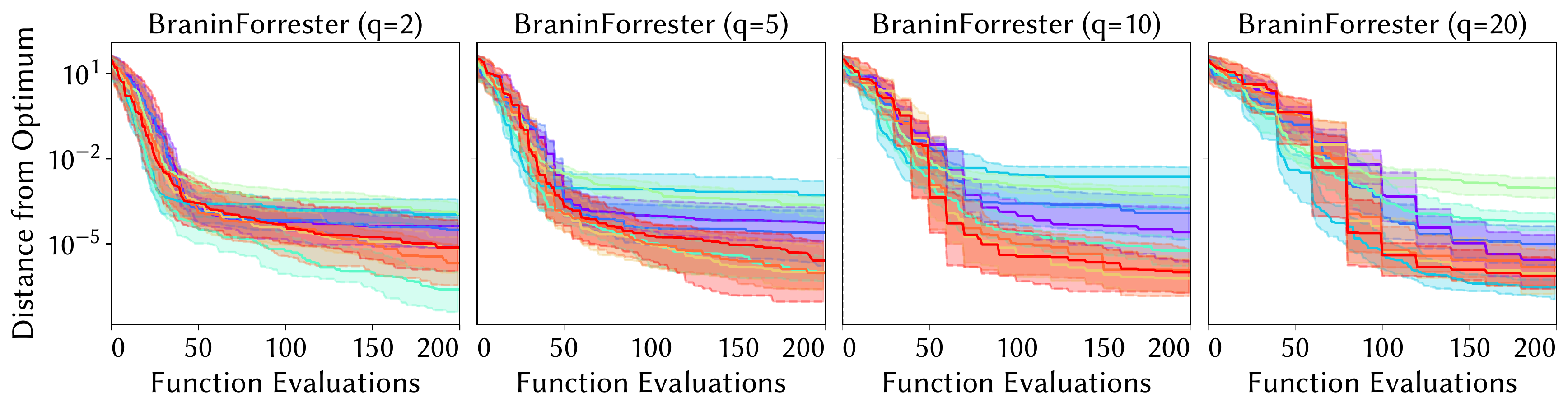}\\
\includegraphics[width=\columnwidth, clip, trim={0 0 0 0}]{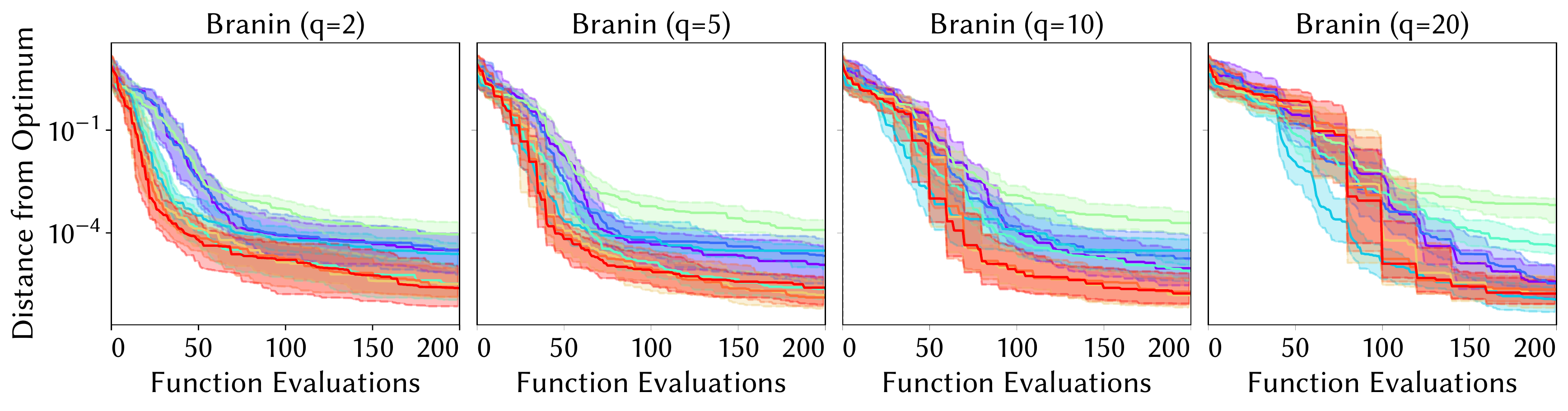}\\
\includegraphics[width=\columnwidth, clip, trim={0 0 0 0}]{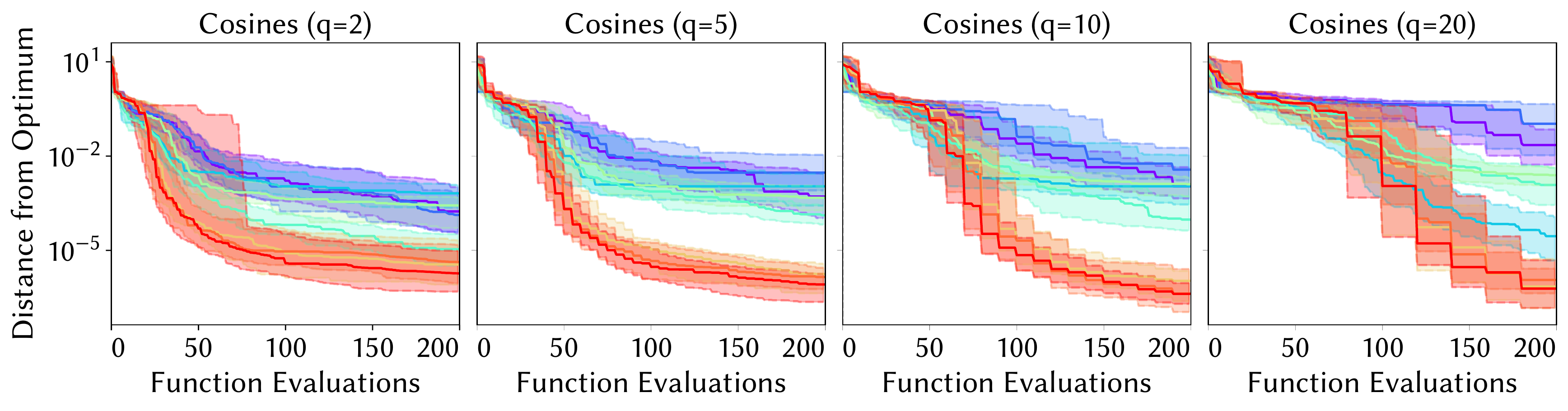}\\
\includegraphics[width=0.65\columnwidth, clip, trim={10 10 10 10}]{figs/convergence_LEGEND}%
\caption{Illustrative convergence plots for the  WangFreitas ($d=1$), 
BraninForrester ($d=2$), Branin ($d=2$) and Cosines ($d=2$) benchmark problems
(rows) for four batch sizes $q \in \{5,10,20\}$ (columns). Each plot shows the
median difference between the best function value seen and the true optimum, 
with shading representing the interquartile range across the 51 runs.}
\label{fig:conv_plots_1}
\end{figure}

\begin{figure}[ht!]
\includegraphics[width=\columnwidth, clip, trim={0 0 0 0}]{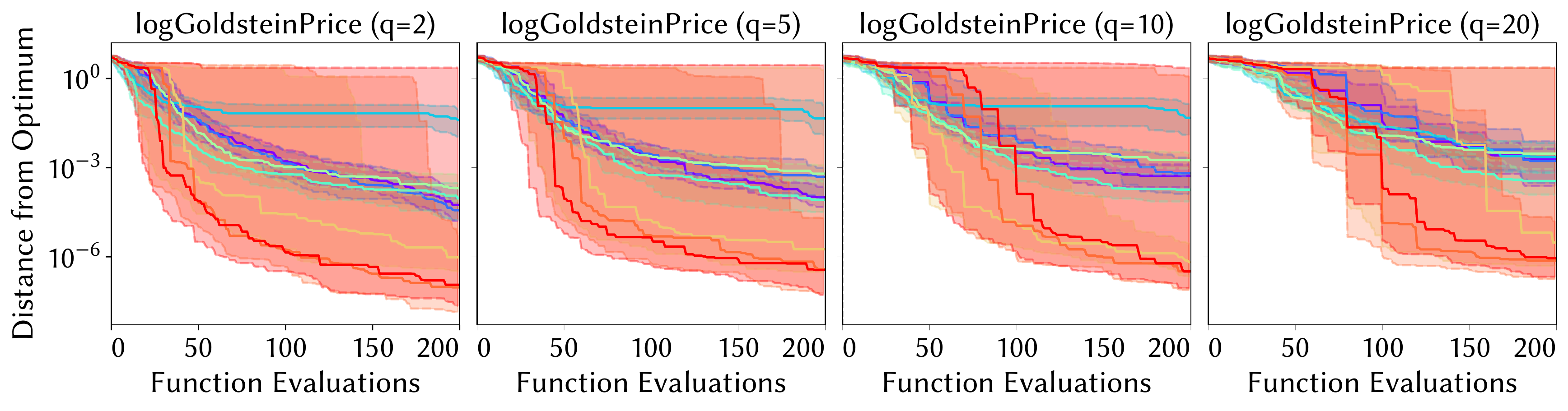}\\
\includegraphics[width=\columnwidth, clip, trim={0 0 0 0}]{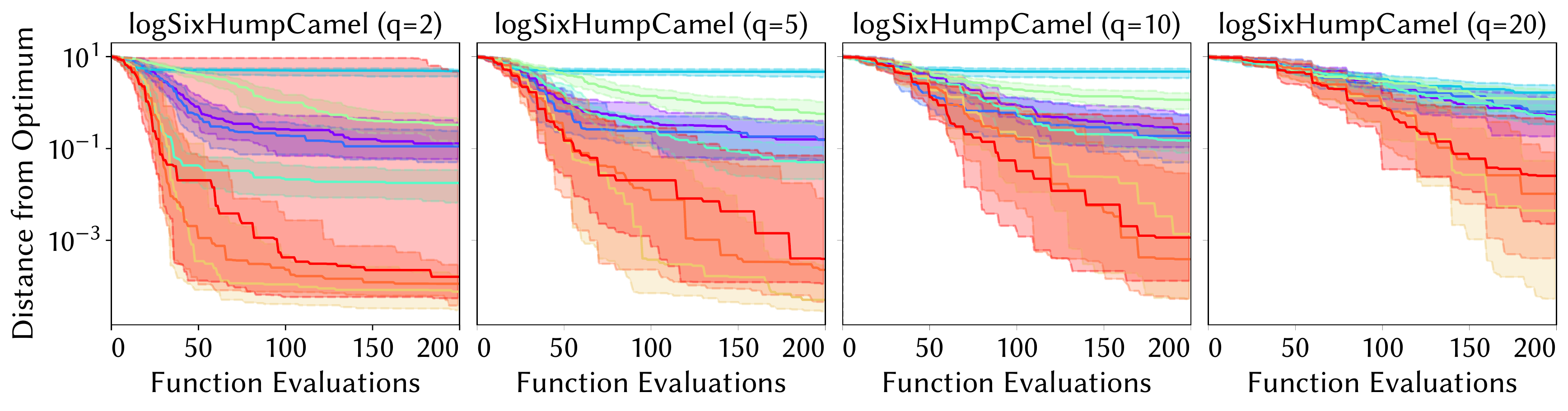}\\
\includegraphics[width=\columnwidth, clip, trim={0 0 0 0}]{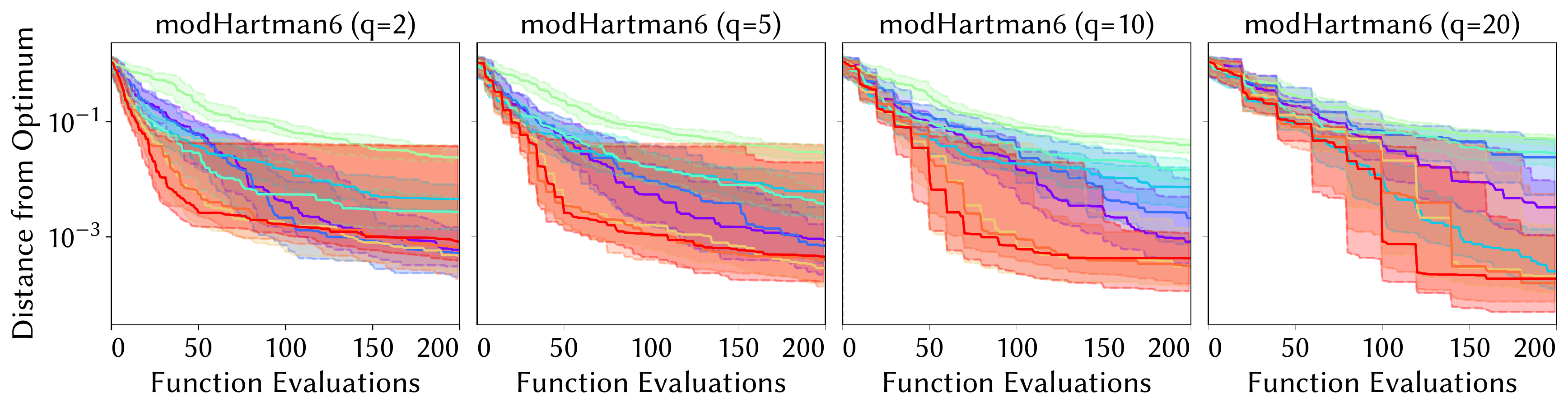}\\
\includegraphics[width=\columnwidth, clip, trim={0 0 0 0}]{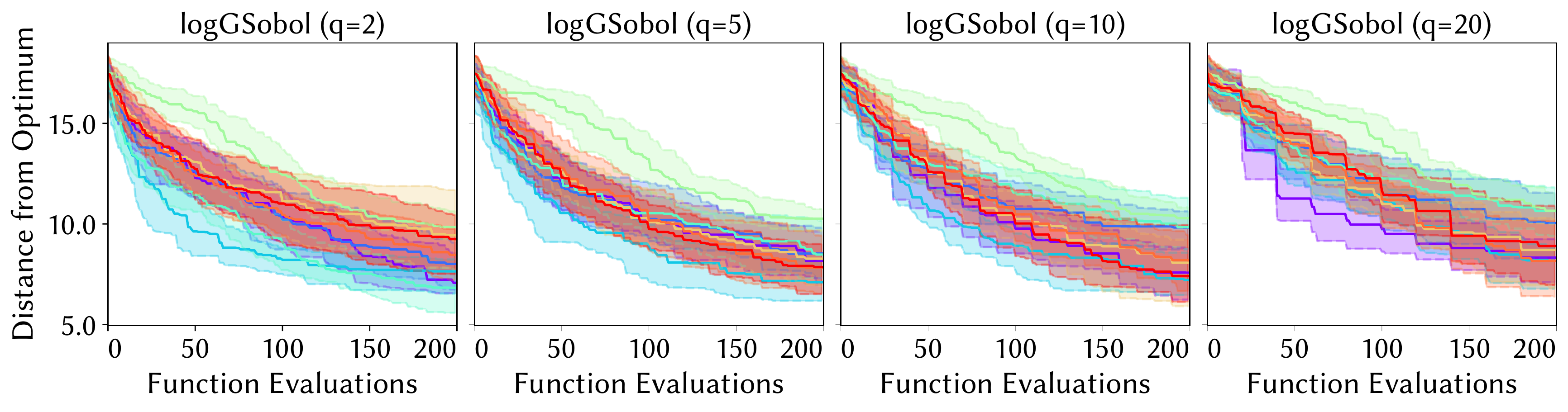}\\
\includegraphics[width=0.65\columnwidth, clip, trim={10 10 10 10}]{figs/convergence_LEGEND}%
\caption{Illustrative convergence plots for the logGoldsteinPrice ($d=1$), 
logSixHumpCamel ($d=2$), modHartman6 ($d=6$) and logGSobol ($d=10$) benchmark
problems (rows) for four batch sizes $q \in \{5,10,20\}$ (columns). Each plot
shows the median difference between the best function value seen and the true
optimum, with shading representing the interquartile range across the 51 runs.}
\label{fig:conv_plots_2}
\end{figure}

\begin{figure}[ht!]
\includegraphics[width=\columnwidth, clip, trim={0 0 0 0}]{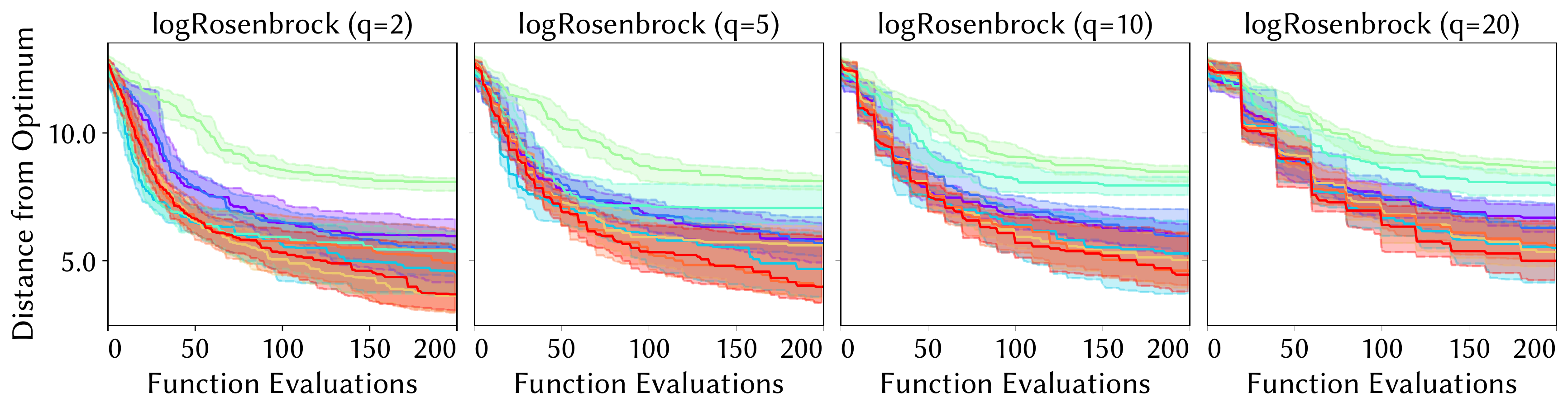}\\
\includegraphics[width=\columnwidth, clip, trim={0 0 0 0}]{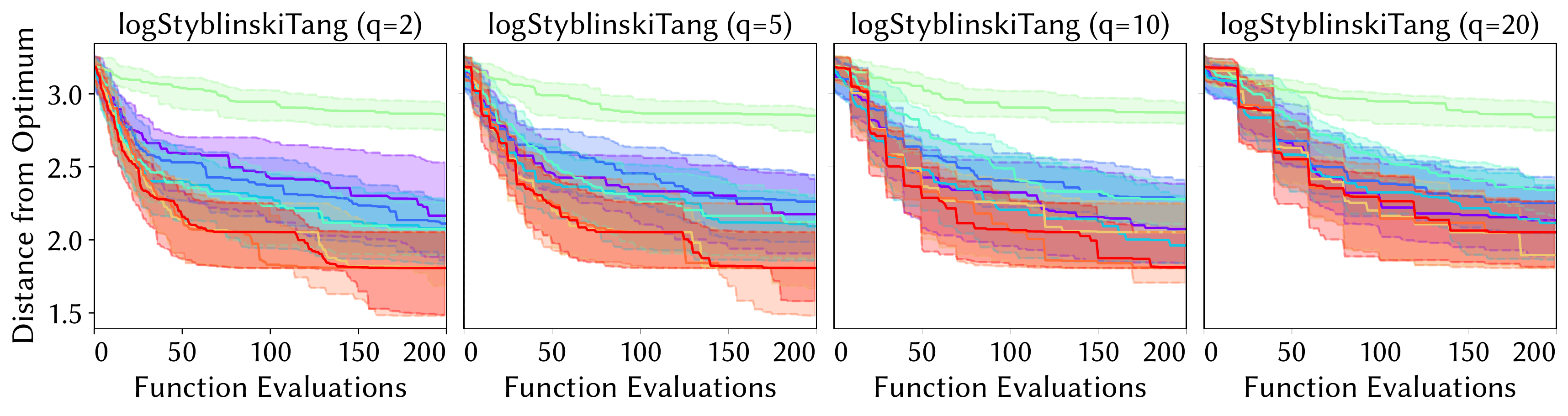}\\
\includegraphics[width=0.65\columnwidth, clip, trim={10 10 10 10}]{figs/convergence_LEGEND}%
\caption{Illustrative convergence plots for the logRosenbrock ($d=10$)
and logStyblinskiTang ($d=10$) benchmark problems (rows) for four batch sizes 
$q \in \{5,10,20\}$ (columns). Each plot shows the median difference between 
the best function value seen and the true optimum, with shading representing 
the interquartile range across the 51 runs.}
\label{fig:conv_plots_3}
\end{figure}

\clearpage
\subsection{Active Learning for Robot Pushing}
\label{sec:robot_pushing}
  \begin{table}[ht!]
  \setlength{\tabcolsep}{2pt}
  \sisetup{table-format=1.2e-1,table-number-alignment=center}
  \resizebox{1\textwidth}{!}{%
  \begin{tabular}{l | SS| SS| SS| SS}
    \toprule
    \bfseries Method
    & \multicolumn{2}{c|}{\bfseries \pushfour (q=2)}
    & \multicolumn{2}{c|}{\bfseries \pushfour (q=5)} 
    & \multicolumn{2}{c|}{\bfseries \pushfour (q=10)}
    & \multicolumn{2}{c}{\bfseries \pushfour (q=20)} \\ 
    & \multicolumn{1}{c}{Median} & \multicolumn{1}{c|}{MAD}
    & \multicolumn{1}{c}{Median} & \multicolumn{1}{c|}{MAD}
    & \multicolumn{1}{c}{Median} & \multicolumn{1}{c|}{MAD}
    & \multicolumn{1}{c}{Median} & \multicolumn{1}{c}{MAD}  \\ \midrule
    LP & 1.78e-01 & 1.30e-01
       & 1.84e-01 & 1.14e-01
       & 1.71e-01 & 1.27e-01
       & 2.32e-01 & 1.87e-01 \\
    PLAyBOOK & 1.97e-01 & 1.19e-01
             & 1.66e-01 & 1.48e-01
             & 1.36e-01 & 1.19e-01
             & 2.18e-01 & 2.49e-01 \\
    KB & 1.50e-01 & 8.05e-02 
       & 1.52e-01 & 1.05e-01
       & 1.97e-01 & 1.18e-01
       & \statsimilar 1.53e-01 & \statsimilar 9.83e-02 \\
    qEI & 1.33e-01 & 8.07e-02
        & 2.00e-01 & 1.09e-01
        & 2.29e-01 & 1.22e-01
        & 2.54e-01 & 1.42e-01 \\
    TS & 2.89e-01 & 1.77e-01
       & 3.05e-01 & 2.08e-01
       & 3.38e-01 & 2.03e-01
       & 3.39e-01 & 1.70e-01 \\
    \eSRS (0.1) & \best 1.02e-02 & \best 9.70e-03
                & \statsimilar 3.11e-02 & \statsimilar 2.95e-02
                & \statsimilar 5.53e-02 & \statsimilar 4.84e-02
                & \statsimilar 1.55e-01 & \statsimilar 1.38e-01 \\
    \eSPF (0.1) & \statsimilar 1.71e-02 & \statsimilar 1.27e-02
                & \statsimilar 2.73e-02 & \statsimilar 2.73e-02
                & \best 4.65e-02 & \best 4.36e-02
                & \statsimilar 1.37e-01 & \statsimilar 1.31e-01 \\
    \eExploit & \statsimilar 1.05e-02 & \statsimilar 1.03e-02
              & \best 2.30e-02 & \best 2.03e-02
              & \statsimilar 5.44e-02 & \statsimilar 5.80e-02
              & \best 1.02e-01 & \best 9.97e-02 \\
\bottomrule
    \toprule
    \bfseries Method
    & \multicolumn{2}{c|}{\bfseries \pusheight (q=2)}
    & \multicolumn{2}{c|}{\bfseries \pusheight (q=5)}
    & \multicolumn{2}{c|}{\bfseries \pusheight (q=10)}
    & \multicolumn{2}{c}{\bfseries \pusheight (q=20)} \\ 
    & \multicolumn{1}{c}{Median} & \multicolumn{1}{c|}{MAD}
    & \multicolumn{1}{c}{Median} & \multicolumn{1}{c|}{MAD}
    & \multicolumn{1}{c}{Median} & \multicolumn{1}{c|}{MAD}
    & \multicolumn{1}{c}{Median} & \multicolumn{1}{c}{MAD}  \\ \midrule
    LP & \statsimilar 2.28e+00 & \statsimilar 1.43e+00
       & \statsimilar 2.06e+00 & \statsimilar 1.46e+00
       & \statsimilar 2.32e+00 & \statsimilar 1.55e+00
       & 2.71e+00 & 1.86e+00 \\
    PLAyBOOK & \statsimilar 1.92e+00 & \statsimilar 1.58e+00
             & \statsimilar 2.42e+00 & \statsimilar 1.58e+00
             & \statsimilar 2.60e+00 & \statsimilar 1.73e+00
             & 2.85e+00 & 1.62e+00 \\
    KB & \statsimilar 2.46e+00 & \statsimilar 1.31e+00
       & 2.38e+00 & 1.91e+00
       & \statsimilar 2.05e+00 & \statsimilar 1.44e+00
       & \statsimilar 2.17e+00 & \statsimilar 1.51e+00 \\
    qEI & \statsimilar 1.82e+00 & \statsimilar 1.41e+00
        & \statsimilar 2.25e+00 & \statsimilar 1.65e+00
        & \statsimilar 2.24e+00 & \statsimilar 1.20e+00
        & \statsimilar 1.97e+00 & \statsimilar 1.45e+00 \\
    TS & 2.96e+00 & 1.61e+00
       & 3.18e+00 & 1.82e+00
       & \statsimilar 2.51e+00 & \statsimilar 1.72e+00
       & 3.05e+00 & 2.27e+00 \\
    \eSRS (0.1) & \best 1.50e+00 & \best 2.08e+00
                & \best 1.82e+00 & \best 1.27e+00
                & \statsimilar 2.26e+00 & \statsimilar 2.00e+00
                & \statsimilar 1.85e+00 & \statsimilar 1.58e+00 \\
    \eSPF (0.1) & \statsimilar 1.86e+00 & \statsimilar 1.54e+00
                & \statsimilar 2.00e+00 & \statsimilar 1.34e+00
                & \best 1.65e+00 & \best 1.51e+00
                & \best 1.76e+00 & \best 1.46e+00 \\
    \eExploit & \statsimilar 1.90e+00 & \statsimilar 1.86e+00
              & \statsimilar 2.00e+00 & \statsimilar 1.75e+00
              & \statsimilar 2.05e+00 & \statsimilar 1.73e+00
              & \statsimilar 2.24e+00 & \statsimilar 2.02e+00 \\
\bottomrule
  \end{tabular}%
  }%
\caption{Optimisation results for the robot pushing problems with batch sizes
$q \in \{2,5,10,20\}$. Median absolute distance from the optimum (left) and 
median absolute deviation from the median (MAD, right) after 200 function 
evaluations across the 51 runs. The method with the lowest median performance 
is shown in dark grey, with those with statistically equivalent performance 
shown in light grey.}
  \label{tbl:results_push_all}
  \end{table}
\begin{figure}[ht!]
\includegraphics[width=\columnwidth, clip, trim={0 0 0 0}]{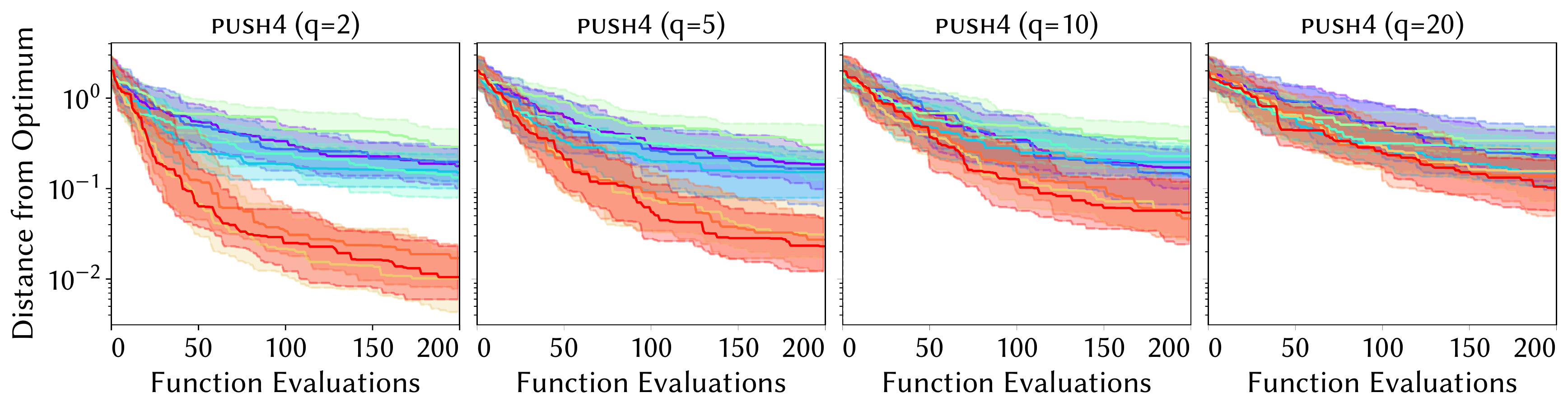}\\
\includegraphics[width=\columnwidth, clip, trim={0 0 0 0}]{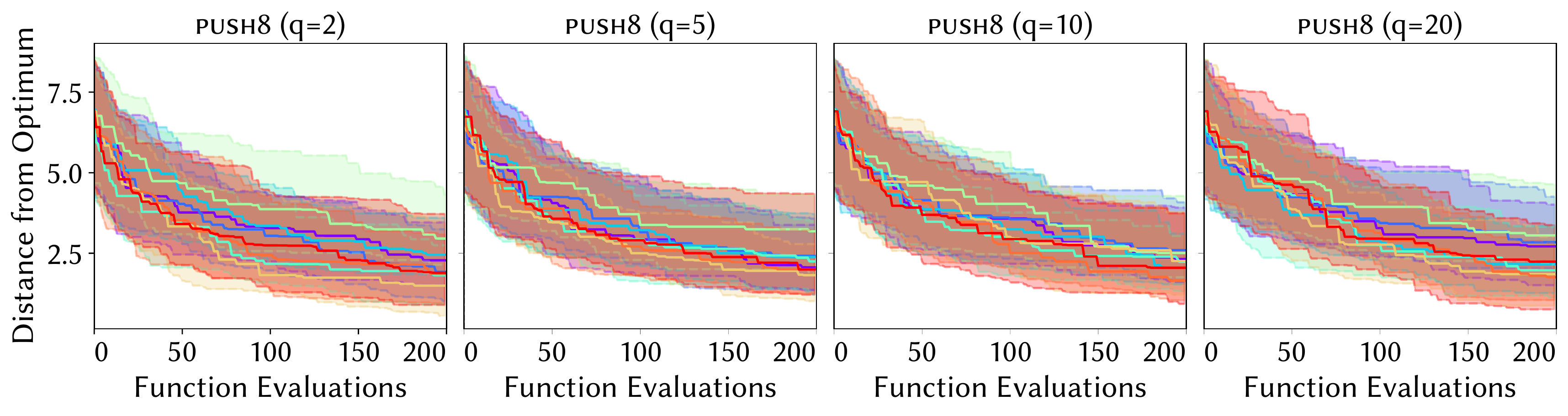}\\
\includegraphics[width=0.65\columnwidth, clip, trim={10 10 10 10}]{figs/convergence_LEGEND}%
\caption{Illustrative convergence plots for the \pushfour ($d=4$) and
\pusheight ($d=8$) real-world robot pushing active learning functions (rows)
for four batch sizes $q \in \{5,10,20\}$ (columns). Each plot shows the median 
difference between the best function value seen and the true optimum, with 
shading representing the interquartile range across the 51 runs.}
\label{fig:conv_plots_all_push}
\end{figure}

\clearpage
\subsection{Pipe Shape Optimisation}
\label{sec:pitzdaily}
  \begin{table}[ht!]
  \setlength{\tabcolsep}{2pt}
  \sisetup{table-format=1.2e-1,table-number-alignment=center}
  \resizebox{1\textwidth}{!}{%
  \begin{tabular}{l | SS| SS| SS| SS}
    \toprule
    \bfseries Method
    & \multicolumn{2}{c|}{\bfseries PitzDaily (q=2)} 
    & \multicolumn{2}{c|}{\bfseries PitzDaily (q=5)} 
    & \multicolumn{2}{c|}{\bfseries PitzDaily (q=10)} 
    & \multicolumn{2}{c}{\bfseries PitzDaily (q=20)} \\
    & \multicolumn{1}{c}{Median} & \multicolumn{1}{c|}{MAD}
    & \multicolumn{1}{c}{Median} & \multicolumn{1}{c|}{MAD}
    & \multicolumn{1}{c}{Median} & \multicolumn{1}{c|}{MAD}
    & \multicolumn{1}{c}{Median} & \multicolumn{1}{c}{MAD} \\ \midrule
    LP & \statsimilar 8.67e-02 & \statsimilar 3.38e-03 
       & 8.62e-02 & 3.74e-03 & \statsimilar 8.72e-02 
       & \statsimilar 3.98e-03 & \statsimilar 8.87e-02 
       & \statsimilar 7.02e-03 \\
    PLAyBOOK & \statsimilar 8.65e-02 & \statsimilar 3.80e-03
             & \statsimilar 8.60e-02 & \statsimilar 3.15e-03
             & \statsimilar 8.79e-02 & \statsimilar 5.21e-03
             & 9.03e-02 & 5.56e-03 \\
    KB & \statsimilar 8.44e-02 & \statsimilar 1.58e-03
       & \best 8.45e-02 & \best 2.23e-03
       & \best 8.55e-02 & \best 2.08e-03
       & \statsimilar 8.49e-02 & \statsimilar 2.44e-03 \\
    qEI & \statsimilar 8.43e-02 & \statsimilar 1.49e-03
		& 8.60e-02 & 3.72e-03
		& 9.04e-02 & 4.88e-03
		& 9.51e-02 & 4.94e-03 \\
    TS & 9.44e-02 & 5.25e-03 
       & 9.31e-02 & 4.19e-03
       & 9.32e-02 & 6.13e-03
       & 9.30e-02 & 3.89e-03 \\
    \eSRS (0.1) & \statsimilar 8.40e-02 & \statsimilar 1.49e-03
                & \statsimilar 8.49e-02 & \statsimilar 2.65e-03
                & \statsimilar 8.58e-02 & \statsimilar 4.03e-03
                & \statsimilar 8.55e-02 & \statsimilar 3.51e-03 \\
    \eSPF (0.1) & \best 8.40e-02 & \best 1.42e-03
                & \statsimilar 8.46e-02 & \statsimilar 2.18e-03
                & \statsimilar 8.57e-02 & \statsimilar 3.81e-03
                & \best 8.49e-02 & \best 2.73e-03 \\
    \eExploit & 8.79e-02 & 5.91e-03
              & 8.83e-02 & 7.26e-03
              & \statsimilar 8.81e-02 & \statsimilar 7.03e-03 
              & \statsimilar 8.84e-02 & \statsimilar 6.53e-03 \\
\bottomrule
  \end{tabular}%
  }%
\caption{Optimisation results for the pipe shape optimisation problem with
batch sizes $q \in \{2,5,10,20\}$. Median absolute distance from the optimum
(left) and median absolute deviation from the median (MAD, right) after 200
function evaluations across the 51 runs. The method with the lowest median
performance is shown in dark grey, with those with statistically equivalent
performance shown in light grey.}
  \label{tbl:results_pitzdaily_all}
  \end{table}
  
\begin{figure}[ht!]
\includegraphics[width=1\columnwidth, clip, trim={0 0 0 0}]{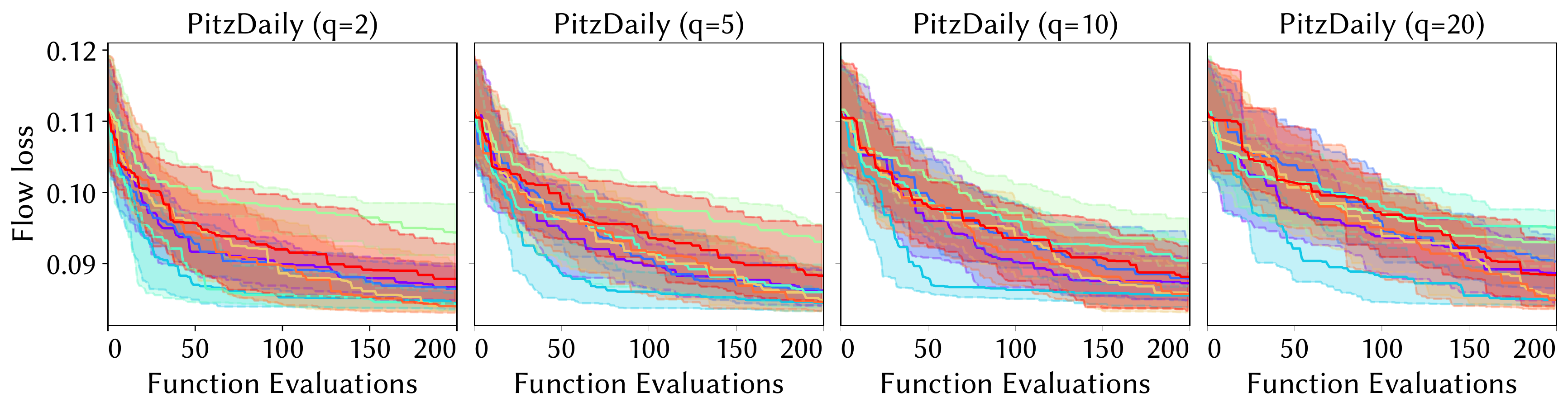}\\
\includegraphics[width=0.65\columnwidth, clip, trim={10 10 10 10}]{figs/convergence_LEGEND}%
\caption{Illustrative convergence plots for the PitzDaily ($d=10$) real-world 
pipe shape optimisation problem for four batch sizes $q \in \{5,10,20\}$
(columns). Each plot shows the median difference between the best function 
value seen and the true optimum, with shading representing the interquartile
range across the 51 runs.}
\label{fig:conv_plots_pitzdaily_supp}
\end{figure}